\documentclass[dvipsnames,format=sigconf,nonacm]{acmart}
\acmConference{}{}{}

\AtBeginDocument{%
  }

\usepackage{subcaption}
\usepackage{booktabs}
\usepackage{xcolor}

\usepackage{tcolorbox}
\usepackage{subcaption}
\usepackage{listings}
\usepackage{float}

\usepackage{multirow} 
\definecolor{codegreen}{rgb}{0,0.6,0}
\definecolor{codegray}{rgb}{0.5,0.5,0.5}
\definecolor{codepurple}{rgb}{0.58,0,0.82}
\definecolor{backcolour}{rgb}{0.95,0.95,0.92}
\definecolor{promptbg}{RGB}{245,247,250} 
\begin{document}
\newtcolorbox[auto counter]{promptbox}[2][]{
  colback=promptbg,
  colframe=black!60,
  boxrule=0.5pt,
  title=Box~\thetcbcounter: #2,#1
}
\title{Evolutionary Discovery of Reinforcement Learning Algorithms via Large Language Models}

\author{Alkis Sygkounas}
\affiliation{
  \institution{Machine Perception and Interaction Lab, Örebro University}
  \country{Sweden}
}
\email{alkis.sygkounas@oru.se}

\author{Amy Loutfi}
\affiliation{
  \institution{Machine Perception and Interaction Lab, Örebro University}
  \country{Sweden}
}
\email{amy.loutfi@oru.se}

\author{Andreas Persson}
\affiliation{
  \institution{Machine Perception and Interaction Lab, Örebro University}
  \country{Sweden}
}
\email{andreas.persson@oru.se}

\begin{abstract}
Reinforcement learning algorithms are defined by their learning update rules, which are typically hand-designed and fixed. We present an evolutionary framework for discovering reinforcement learning algorithms by searching directly over executable update rules that implement complete training procedures. The approach builds on REvolve, an evolutionary system that uses large language models as generative variation operators, and extends it from reward-function discovery to algorithm discovery. To promote the emergence of nonstandard learning rules, the search excludes canonical mechanisms such as actor--critic structures, temporal-difference losses, and value bootstrapping. Because reinforcement learning algorithms are highly sensitive to internal scalar parameters, we introduce a post-evolution refinement stage in which a large language model proposes feasible hyperparameter ranges for each evolved update rule. Evaluated end-to-end by full training runs on multiple Gymnasium benchmarks, the discovered algorithms achieve competitive performance relative to established baselines, including SAC, PPO, DQN, and A2C.
\end{abstract}

\maketitle
\section{Introduction}

In reinforcement learning, the update rule determines how experience is transformed into parameter updates, which directly shape the learning behavior. Although automated discovery has been applied to some aspects of reinforcement learning, such as architecture design \cite{zoph2017nas,pham2018efficient,sato2021advantagenas}, hyperparameter tuning \cite{paul2019fast,xu2018meta,eimer2023hyperparameters}, and training-signal design \cite{he2022reinforcement,zou2021learning}, the update rule typically remains fixed. A small number of works have attempted to further elevate the discovery by exploring the learning of update rules intrinsically, either by optimizing differentiable update functions \cite{oh2020discovering}, or by evolving structured loss expressions \cite{coreyes2021evolving}. However, these methods operate over restricted, parameterized representations of the learning rule, such as differentiable update networks or symbolic loss forms, and therefore cannot generate fundamentally different update rules expressed as executable training logic. As a result, the update rule remains one of the core components of reinforcement learning that has not yet been the subject of fully automated design.

Searching over reinforcement-learning update rules is incompatible with gradient-based or other locally guided optimization because the search space consists of discrete program structures implementing full training procedures, rather than continuous parameters or differentiable update modules. Even small changes to the update logic can induce qualitatively different learning dynamics, and reinforcement-learning algorithms are known to be highly sensitive to implementation details \cite{henderson2018deep, engstrom2020implementation}. Hyperparameters that stabilize one update rule often fail for another \cite{patterson2024empirical}, preventing effective local refinement through small perturbations. As a result, each update rule must be treated as a distinct algorithm whose quality can only be assessed through complete training runs. Under these conditions, population-based evolutionary methods are a natural fit, since they compare candidate algorithms using end-to-end performance without requiring gradients or smooth objective structure \cite{jaderberg2017population, khadka2018evolution, coreyes2021evolving}.

Evolutionary algorithms and genetic programming have been applied to a wide range of machine-learning tasks, including neural network evolution, hyperparameter optimization, controller synthesis, and symbolic policy search \cite{franke2019neural,jaderberg2017population,khadka2018evolution}. A key practical requirement in these settings is the ability to generate offspring that are both syntactically valid and executable, enabling evolutionary operators to explore the search space without producing predominantly invalid variants \cite{coreyes2021evolving}. Reinforcement-learning update rules, however, violate this requirement because they are implemented as tightly coupled training procedures rather than isolated objective functions. As a result, even small structural code changes can break dependencies between components or invalidate required data flows, leading to failed or undefined training behavior \cite{henderson2018deep, engstrom2020implementation, patterson2024empirical}. Consequently, unguided mutation and crossover are unlikely to reliably produce functional update rules, underscoring the need for a guided mechanism that proposes coherent, executable modifications to complete training logic.

Recent work suggests that large language models (LLMs) can serve as such proposal mechanisms within evolutionary pipelines, enabling structured exploration of complex design spaces. In reinforcement learning, this approach has been used to optimize reward functions by generating reward implementations whose utility is assessed through training outcomes \cite{eureka, revolve}. Related ideas have also been explored for program and code evolution within genetic programming frameworks \cite{hemberg2024evolving, coreyes2021evolving}, as well as for broader evolutionary computation, such as population-based and quality-diversity search in neural architecture optimization \cite{nasir2024llmatic}. However, across these lines of work, evolutionary search typically targets individual components of learning systems, while the reinforcement-learning update rule itself remains fixed.

\begin{figure}[h]
    \centering
    \includegraphics[width=0.96\linewidth]{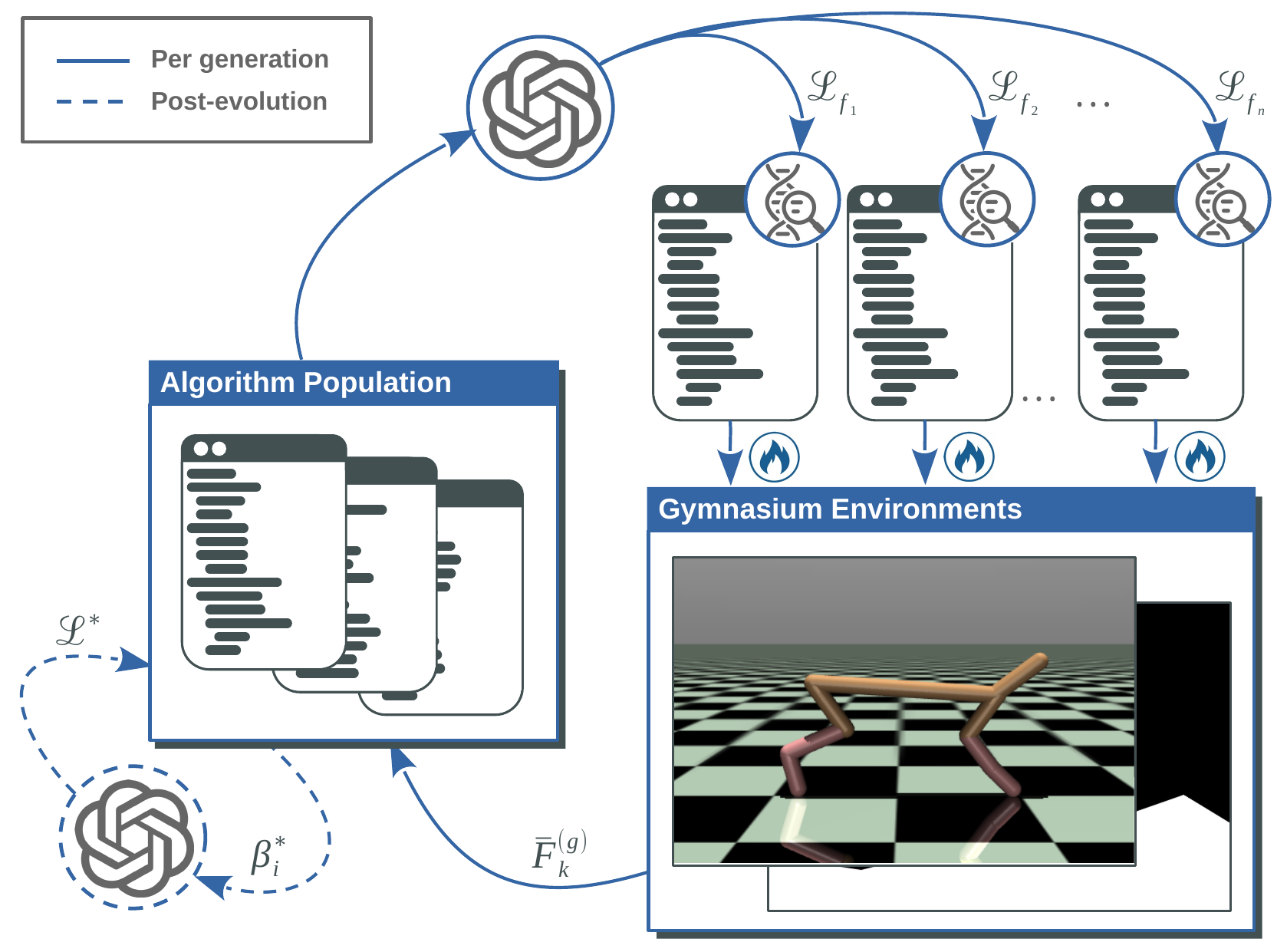}
    \caption{Illustrative overview of the proposed method. A population of candidate algorithms is iteratively evolved. In each generation (solid arrows), a large language model proposes coherent variants ($\mathcal{L}_{f_1}, \mathcal{L}_{f_2}, \ldots, \mathcal{L}_{f_n}$), which are evaluated via training in Gymnasium environments to obtain fitness scores ($\bar{F}^{(g)}_k$) that are subsequently used for selection and population updates. Post-evolution (dashed arrows), a hyperparameter setting ($\beta_i^\star$), selected via LLM-guided optimization, optimizes the resulting best update rule ($\mathcal{L}^\star$), which is then advanced for final evaluation.}
    \label{fig:framework_overview}
\end{figure}

In this work, we investigate evolutionary search with large language models to discover novel reinforcement-learning algorithms by directly searching over learning update rules. Our approach builds on REvolve \cite{revolve}, which formulates reward-function discovery as an evolutionary process using island populations \cite{island_ea_seminal} and language-model-based variation. We apply the same evolutionary structure to a different search space. However, instead of evolving reward functions, we evolve learning update rules as executable code under a fixed policy architecture, optimizer, and training configuration. In this setting, each candidate update rule defines a distinct reinforcement-learning algorithm. To encourage the synthesis of noncanonical learning rules and to assess whether genuinely new update rules can emerge, we prohibit the use of standard mechanisms such as actor–critic decomposition \cite{konda1999actor}, bootstrapped value targets \cite{sutton1999policy}, and temporal-difference error computation \cite{sutton1999policy}. Each candidate algorithm is evaluated by full training on a fixed set of environments, and selection is based on aggregated empirical performance. We further extend REvolve with a post-evolution hyperparameter optimization applied to top-performing algorithms to evaluate robustness and improve stability under fixed training budgets. An overview of the proposed method is illustrated in Figure~\ref{fig:framework_overview}. In summary, the main contributions of this paper are as follows:
\begin{itemize}
    \item We formulate reinforcement-learning algorithm discovery as an evolutionary search process over executable learning update rules, enabling the direct synthesis of complete learning algorithms.
    \vspace{4pt}
    \item We extend REvolve to search in the update-rule space using LLM-guided mutation and crossover, with explicit constraints that exclude canonical RL mechanisms to promote the discovery of nonstandard algorithms.
    \vspace{4pt}
    \item We empirically demonstrate that the proposed framework discovers new reinforcement-learning algorithms that achieve competitive performance across multiple Gymnasium benchmarks after post-evolution hyperparameter optimization.
\end{itemize}

\section{Related Work}

Several lines of work have focused on automating components of reinforcement learning systems that are external to the learning update rule itself. This includes neural architecture search for reinforcement learning agents \cite{zoph2017nas,pham2018efficient,sato2021advantagenas}, large-scale hyperparameter optimization frameworks targeting training stability and sample efficiency \cite{jaderberg2017population,xu2018meta}, and methods that automate the design of auxiliary objectives or training signals \cite{jaderberg2016reinforcement,he2022reinforcement}. These approaches improve performance by modifying architectures, hyperparameters, or training signals, while the underlying update rule is typically held fixed.

Evolutionary and population-based methods provide a complementary line of work that improves learning performance by operating over policies, parameters, or architectures within a fixed training loop. This includes neuro-evolution approaches that evolve policy parameters or network structures \cite{stanley2002evolving,real2019regularized}, evolutionary strategies applied to policy optimization \cite{salimans2017evolution,khadka2018evolution}, population-based training methods that adapt hyperparameters during learning \cite{jaderberg2017population}, and evolutionary generation of environments and curricula \cite{wang2019paired}. More recently, large language models have been incorporated into similar iterative optimization pipelines for reinforcement learning and related domains, including generating or refining reward specifications and training objectives \cite{eureka,revolve}, and proposing policy-level code or heuristics that are evaluated and selected through performance \cite{hemberg2024evolving}.

Collectively, existing approaches have explored automated and evolutionary optimization of architectures, hyperparameters, rewards, environments, and policy parameters. They have additionally introduced language models as generators within such search processes. However, with the exception of work that learns or evolves update rules in restricted forms \cite{oh2020discovering,coreyes2021evolving,andrychowicz2016learning}, these approaches treat the reinforcement learning update rule itself as fixed. In contrast, our work targets algorithm discovery at the level of the update rule, treating the executable learning logic as the object of evolutionary search.

\section{Methodology}

For this work, we consider reinforcement learning problems defined by a Markov decision process $\mathcal{M}=(\mathcal{S},\mathcal{A},P,r)$. At interaction step $t$, the agent observes $s_t \in \mathcal{S}$, selects an action $a_t \in \mathcal{A}$ according to a policy $\pi_\theta(a \mid s)$ and parameterized by $\theta$, receives reward $r_t$, and transitions to $s_{t+1}$, producing experience tuples $e_t = (s_t, a_t, r_t, s_{t+1}, d_t)$. The accumulated experience is denoted $\mathcal{D}_t = \{e_0, \dots, e_t\}$. Hence, a reinforcement learning algorithm is defined by its \emph{learning update rule}, where each candidate update rule $f$ specifies an executable training procedure whose core component is a differentiable loss function:
\[
\mathcal{L}_f(\theta,\xi_t;\mathcal{D}_t),
\]
where $\theta$ denotes the policy parameters (including a fixed architecture trunk and output head), and $\xi_t$ denotes all auxiliary trainable components and internal state introduced by the update rule (e.g., parameters of auxiliary networks or target models) that are updated during training. The induced parameter updates are given by:
\[
\Delta\theta_t = -\,\eta_f \nabla_\theta \mathcal{L}_f(\theta_t,\xi_t;\mathcal{D}_t),
\qquad
\Delta\xi_t = -\,\eta_f^\xi \nabla_{\xi} \mathcal{L}_f(\theta_t,\xi_t;\mathcal{D}_t),
\]
where $\eta_f$ and $\eta_f^\xi$ are rule-specific update coefficients.
The resulting training dynamics are subsequently:
\[
\theta_{t+1} = \theta_t + \Delta\theta_t,
\qquad
\xi_{t+1} = \xi_t + \Delta\xi_t.
\]

The update rule therefore defines the program-level mapping $f(\theta_t,\xi_t,\mathcal{D}_t) := \Delta\theta_t$.
The search space $\mathcal{F}$ consists of all executable update rules of this form that induce valid parameter updates, i.e., $\theta_t + \Delta\theta_t \in \Theta$ for all admissible $(\theta_t,\xi_t,\mathcal{D}_t)$. Under a fixed policy architecture, optimizer, and training loop, each distinct $f \in \mathcal{F}$ defines a distinct reinforcement learning algorithm.

\subsection{Training Performance and Fitness}
Each update rule $f \in \mathcal{F}$ is evaluated by executing a complete reinforcement learning training run. For a fixed environment $\mathcal{M}_i$, let:
\[
\mathcal{T}(f, \mathcal{M}_i) = \{ R_{i,t}(f) \}_{t=1}^{T},
\]
denote the stochastic training-and-evaluation process that produces a sequence of evaluation returns by periodically evaluating the policy $\pi_{\theta_t}$ during training. Training is performed independently on a fixed set of environments $\{\mathcal{M}_i\}_{i=1}^{N}$.

For each environment, performance is summarized by the maximum evaluation return attained over the training horizon,
\[
\mathrm{MTS}_i(f) = \max_{t \le T} R_{i,t}(f),
\]
corresponding to the best policy produced by the update rule during training. To account for differences in reward scale across environments, these scores are normalized using environment-specific reference bounds $L_i$ and $U_i$,
\[
\tilde{F}_i(f) = \frac{\mathrm{MTS}_i(f) - L_i}{U_i - L_i}.
\]
The evolutionary fitness is defined as the mean normalized performance across
environments,
\[
F(f) = \frac{1}{N} \sum_{i=1}^{N} \tilde{F}_i(f).
\]

\subsection{Evolutionary Selection Mechanism}\label{sec:method_fitness}
The evolutionary process follows the same process as presented in REvolve~\cite{revolve}.  
At initialization, each island is populated with an independently sampled set of update rules.
Let $\mathcal{P}^{(g)}_k$ denote the population of island $k$ at generation $g$, with fixed population size $N$.
At each generation, a total of $M$ new candidate update rules are generated across all islands, with variation applied independently within each island population.
Let:
\[
\bar{F}^{(g)}_k=\frac{1}{N}\sum_{f\in\mathcal{P}^{(g)}_k}F(f),
\]
denote the mean fitness of the current population of island $k$.
A newly generated candidate $f$ is accepted if:
\[
F(f) \ge \bar{F}^{(g)}_k.
\]
Accepted candidates replace the lowest-fitness members of $\mathcal{P}^{(g)}_k$, keeping the population size constant.

\subsection{Variation Operators}  \label{eq:levenstein}
New candidates are generated using macro mutation and diversity-aware crossover, selected stochastically with probabilities $p$ and $1-p$, respectively. In REvolve, micro mutations operate on short-reward expressions, where local edits can produce graded behavioral changes. For update rules, early generations contain low-performing algorithms, whose deficiencies are structural rather than local. Hence, minor token edits do not lead to meaningful improvements and often destabilize training. Macro mutation, on the other hand, rewrites exactly one semantically coherent component of the update rule in a single step, enabling substantial changes to the learning logic \cite{real2017large,Koza1992,stanley2002evolving}.

Crossover combines two parent update rules. If parents are selected purely by fitness, crossover frequently recombines a rule with a near duplicate created in a recent mutation, yielding offspring that differ only trivially from their parents and collapsing population diversity \cite{lehman2011evolving}. To avoid this, parent selection incorporates structural dissimilarity measured via normalized Levenshtein distance \cite{lcvenshtcin1966binary}.

Parent~1 is drawn from the current island population $\mathcal{P}^{(g)}_k$ using a fitness-proportional softmax:
\[
P(f_1)=\frac{\exp(\tau\,F(f_1))}{\sum_{f\in\mathcal{P}^{(g)}_k}\exp(\tau\,F(f))},
\qquad \tau>0.
\]

The fitness values are normalized such that $F(f)\in[0,1]$, where the structural dissimilarity measure $d_{\mathrm{lev}}(f_1,f_2)$ is defined as a normalized Levenshtein distance taking values in $[0,1]$.

Conditional on $f_1$, Parent~2 is sampled according to the combined score:
\[
S(f_2\mid f_1)=\alpha F(f_2)+(1-\alpha)\, d_{\mathrm{lev}}(f_1,f_2),
\]
where $d_{\mathrm{lev}}(f_1,f_2)$ is computed between the source-code representations of the corresponding \texttt{compute\_loss} functions, and $\alpha\in[0,1]$ balances fitness and structural dissimilarity. The resulting sampling distribution is consequently:
\[
P(f_2\mid f_1)=
\frac{\exp(\tau\,S(f_2\mid f_1))}
{\sum_{f\in\mathcal{P}^{(g)}_k}\exp(\tau\,S(f\mid f_1))}.
\]

Higher fitness increases selection probability, but structurally similar candidates are penalized unless they offer clear performance advantages. This discourages crossover between a parent and its minor variants and promotes recombination between high-performing yet
distinct update rules. The variation operator then selects macro mutation or crossover with probabilities $p$ and $1-p$, respectively, and applies the selected operator to generate a new candidate.

\subsection{LLM-Based Generative Operator}
Macro mutation and crossover are implemented through a conditional generative operator:
\[
f' \sim q_\phi(f_1,f_2,\text{op},\mathcal{R},\mathcal{E}),
\]
realized using a large language model. The operator takes as input the parent update rules $(f_1,f_2)$, the selected variation
instruction $\text{op}\in\{\text{macro},\text{crossover}\}$, summary training metrics $\mathcal{R}$, and the environment interface $\mathcal{E}$ specifying observation and action spaces. The output $f'$ is represented as an executable implementation of an update rule, including the definition of the loss $\mathcal{L}_{f'}$ and any auxiliary state updates for $\xi$. During evolutionary variation, the generation process is constrained to forbid explicit instantiation of canonical reinforcement learning mechanisms, including actor--critic decomposition, bootstrapped value targets, and temporal-difference updates. These constraints are applied only during evolutionary search.

\subsection{Post-Evolution: LLM-Guided Hyperparameter Optimization (LLM-HPO)}
REvolve terminates after evolutionary search without additional tuning of selected individuals. For the evolution of update rules, this is insufficient as reinforcement learning algorithms are highly sensitive to internal scalar parameters, and a fixed parameterization can severely underestimate the quality of an update rule \cite{henderson2018deep,engstrom2020implementation}. Instead, we let each evolved update rule $\hat f$ define a family of algorithms parameterized by $\beta \in \mathbb{R}^d$, where $\beta$ collects all internal scalar coefficients of the rule.

After evolutionary convergence, we optimize the top-$K$ update rules $\{\hat f_1,\dots,\hat f_K\}$ by approximating:
\[
\max_{\beta \in \mathcal{B}_{\hat f}} F\!\left(f_{\beta}\right),
\]
where $F(\cdot)$ denotes the aggregated fitness across evaluation environments and $\mathcal{B}_{\hat f} \subset \mathbb{R}^d$ is a rule-specific feasible parameter region. Exhaustive search over $\mathcal{B}_{\hat f}$ is infeasible when fitness must be evaluated through full training runs on multiple environments.

For each $\hat f$, the language model is provided with the full update-rule implementation together with the specifications of the evaluation environments, and returns bounded numeric intervals:
\[
\mathcal{B}_{\hat f}
=
\prod_{j=1}^d [\ell_j, u_j],
\]
one for each internal scalar parameter $\beta_j$. These intervals define an LLM-guided search region that restricts exploration to numerically plausible ranges conditioned on both the algorithm structure and the environment suite.

A sweep then \emph{uniformly samples} parameter vectors $\beta$ from $\mathcal{B}_{\hat f}$, instantiates the corresponding update rule $f_{\beta}$, and evaluates its fitness \emph{separately for each environment} using the same evaluation protocol as during evolution. For each $\hat f_i$, the parameter vector:
\[
\beta_i^\star = \arg\max_{\beta} F\!\left(f_{\beta}\right),
\]
observed during this sampling-based optimization is retained, yielding the final refined algorithm $f_{\beta_i^\star}$. The policy architecture, optimizer, rollout procedure, and training loop remain fixed throughout this stage.

\section{Experiments}

For experiments, we perform evolutionary search using two large language models as generative operators, namely GPT-5.2 and Claude 4.5 Opus. For each language model, we conduct two independent evolutionary runs with different evolutionary random seeds. Each run is executed for $10$ generations, and at every generation, each island produces $24$ new candidate update rules. Variation operators are applied with fixed probabilities $p_{\mathrm{macro}}=0.65$ and $p_{\mathrm{cross}}=0.35$, while diversity-aware crossover is regulated by a Levenshtein distance weight $\alpha=0.5$.

\begin{figure*}[t]
    \centering

    \begin{subfigure}{0.19\textwidth}
        \includegraphics[width=\linewidth]{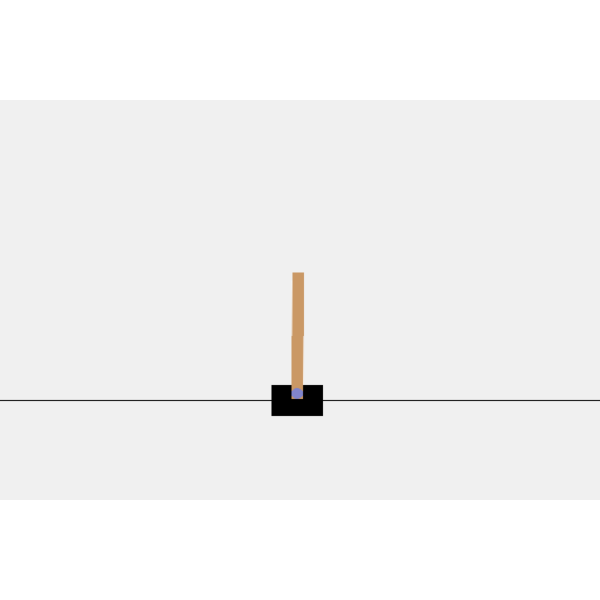}
        \caption{CartPole}
    \end{subfigure}
    \begin{subfigure}{0.19\textwidth}
        \includegraphics[width=\linewidth]{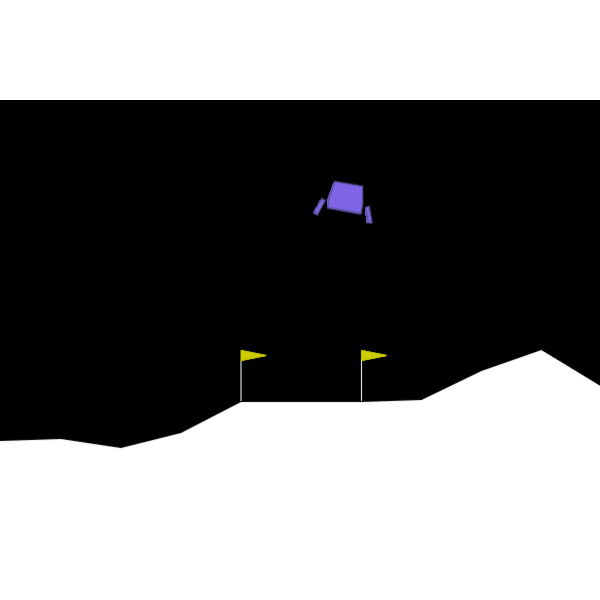}
        \caption{LunarLander}
    \end{subfigure}
    \begin{subfigure}{0.19\textwidth}
        \includegraphics[width=\linewidth]{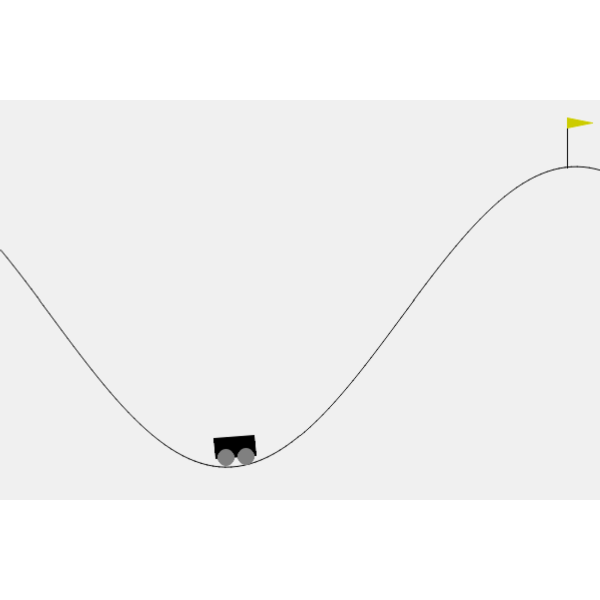}
        \caption{MountainCar}
    \end{subfigure}
    \begin{subfigure}{0.19\textwidth}
        \includegraphics[width=\linewidth]{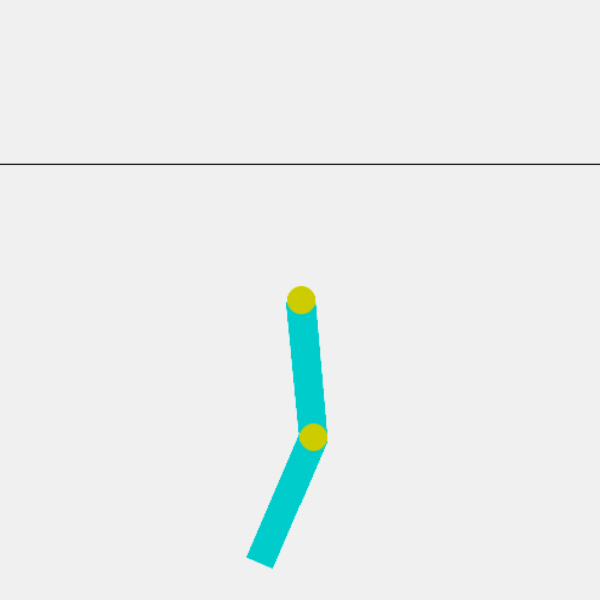}
        \caption{Acrobot}
    \end{subfigure}
    \begin{subfigure}{0.19\textwidth}
        \includegraphics[width=\linewidth]{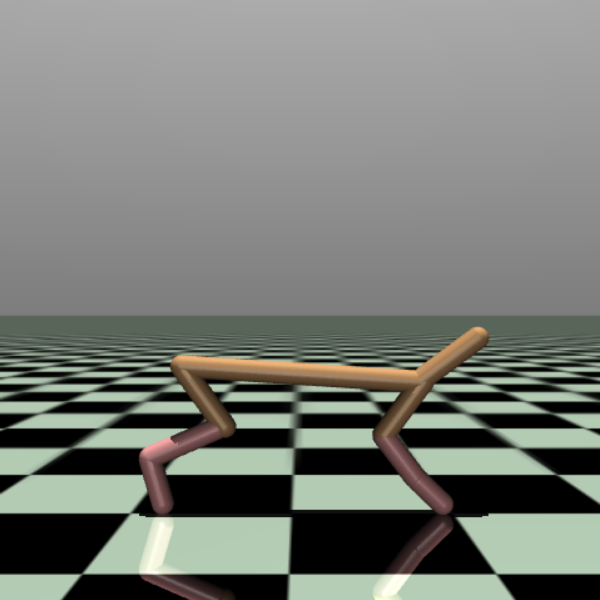}
        \caption{HalfCheetah}
    \end{subfigure}

    \vspace{0.6em}

    \begin{subfigure}{0.19\textwidth}
        \includegraphics[width=\linewidth]{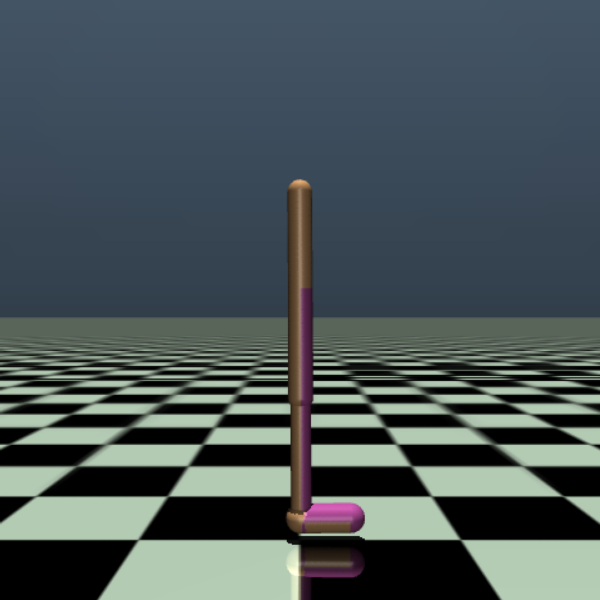}
        \caption{Walker2d}
    \end{subfigure}
    \begin{subfigure}{0.19\textwidth}
        \includegraphics[width=\linewidth]{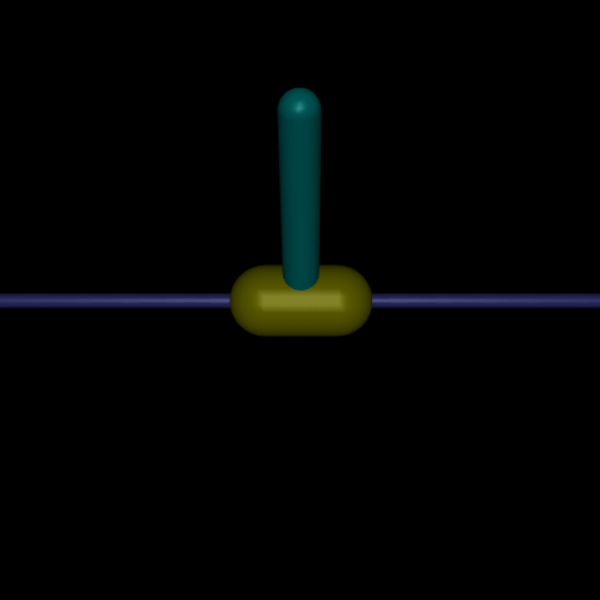}
        \caption{InvertedPendulum}
    \end{subfigure}
    \begin{subfigure}{0.19\textwidth}
        \includegraphics[width=\linewidth]{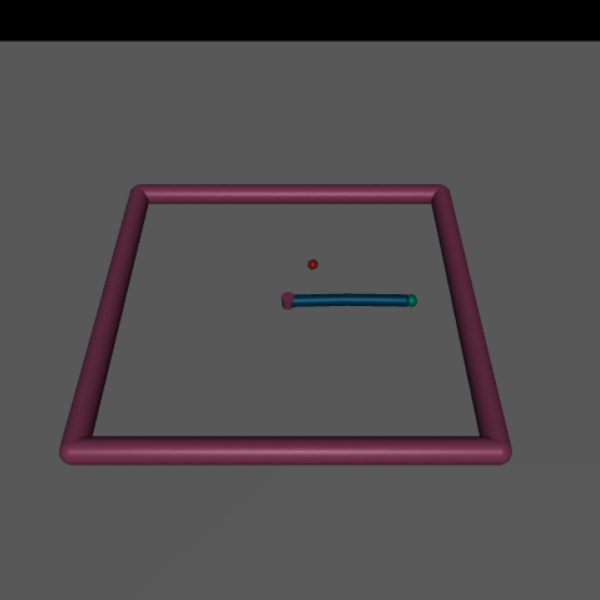}
        \caption{Reacher}
    \end{subfigure}
    \begin{subfigure}{0.19\textwidth}
        \includegraphics[width=\linewidth]{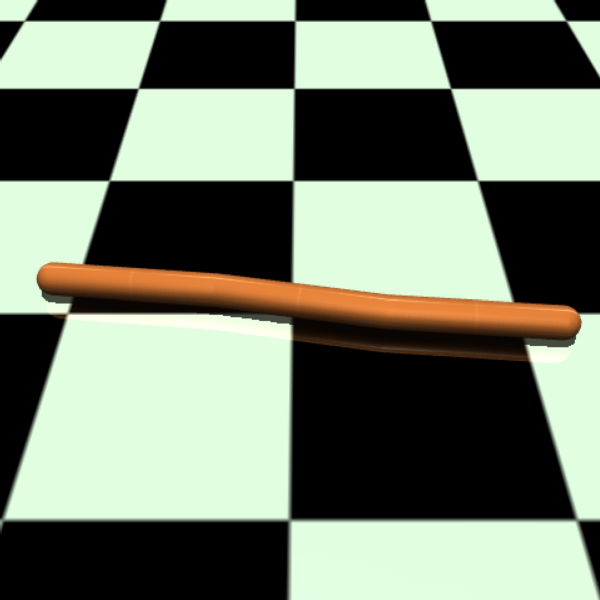}
        \caption{Swimmer} 
    \end{subfigure}
    \begin{subfigure}{0.19\textwidth}
        \includegraphics[width=\linewidth]{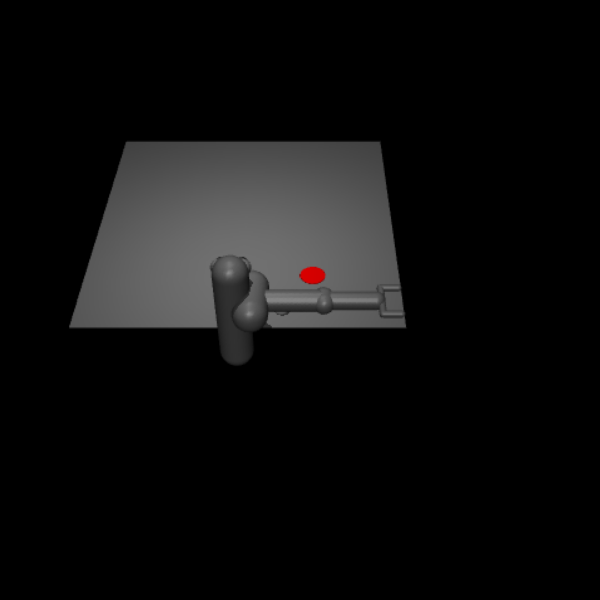}
        \caption{Pusher}
    \end{subfigure}

    \caption{Representative Gymnasium environments used for training (top row) and evaluation (bottom row).}
    \label{fig:env_comparison}
\end{figure*}

Candidate update rules are evaluated across a fixed set of environments: CartPole-v1, MountainCar-v0, Acrobot-v1, LunarLander-v3, and HalfCheetah-v5. CartPole-v1 serves as a minimal control task for verifying basic learning functionality. MountainCar-v0 and Acrobot-v1 feature sparse rewards in discrete action spaces. LunarLander-v3 combines discrete control with shaped rewards and more complex dynamics. HalfCheetah-v5 represents a high-dimensional continuous-control task with dense rewards. Together, this suite spans discrete and continuous action spaces as well as sparse and dense reward regimes, and consists of environments with established reward scales that support consistent fitness normalization and comparison across tasks \cite{gymnasium_general,gym_leaderboard_lunar_lander,gym_mountaincar_wiki,halfcheetah_benchmark_gac}. Figure~\ref{fig:env_comparison} provides representative visualizations of the Gymnasium benchmark environments used for training and evaluation.

Each candidate update rule is evaluated by training it on each of the five environments using $S=5$ independent random seeds per environment, yielding $25$ training runs per candidate. For a given environment, performance is summarized as the average across seeds of the maximum evaluation (best checkpoint model) return achieved during training. These environment-level scores are then normalized to the unit interval using environment-specific reward bounds as described in Section~\ref{sec:method_fitness}, and aggregated across environments to obtain the final scalar fitness used for evolutionary selection. Additional details on environment choice and fitness computation are provided in Appendix~\ref{app:env details}. All evolutionary experiments were conducted on four A100 GPUs with 40\,GB of memory each. Under this setup, a single evolutionary generation required approximately 30 hours of wall-clock time.

\section{Results}

\begin{figure}[t]
    \centering
    \includegraphics[width=0.86\linewidth]{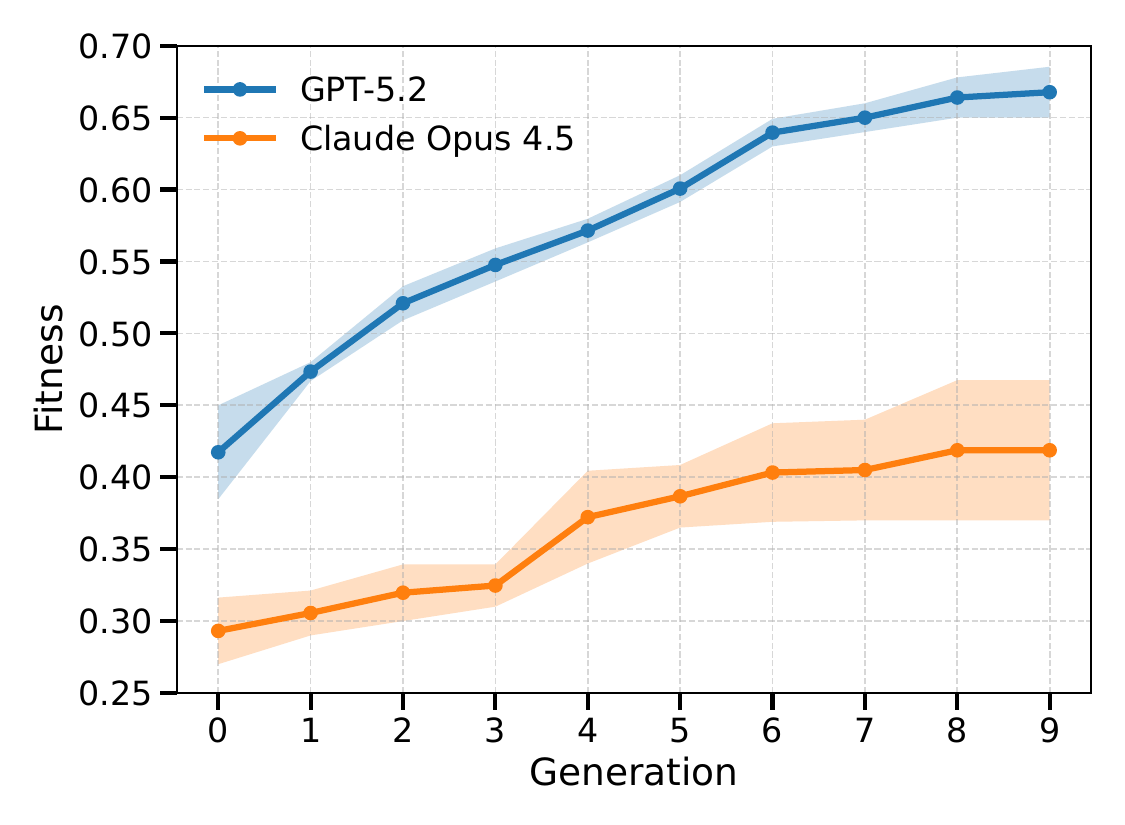}
    \caption{Evolution of maximum population fitness across generations for GPT-5.2 and Claude~4.5~Opus. Curves show the mean across two evolutionary seeds, with shaded regions indicating the standard deviation across seeds.}
    \label{fig:evolution_curves}
\end{figure}

Figure~\ref{fig:evolution_curves} shows the evolutionary progress obtained with GPT-5.2 and Claude~4.5~Opus. For each language model, results are reported for two independent evolutionary runs with different random seeds. At each generation, the plotted value corresponds to the best-performing update rule in the population at that generation, as measured by the aggregated fitness defined in Section~\ref{sec:method_fitness}. Both language models exhibit a monotonic increase in maximum fitness across generations for both evolutionary seeds, as seen in Figure~\ref{fig:evolution_curves}. GPT-5.2 consistently attains higher fitness values than Claude~4.5~Opus throughout evolution, with final-generation fitness in the range $[0.65, 0.69]$ across seeds, compared to $[0.37, 0.47]$ for Claude~4.5~Opus. The shaded regions denote the min--max range across the two evolutionary seeds, indicating moderate variability without qualitative divergence between runs.

\paragraph{Final algorithm selection and evaluation.}
At the end of evolutionary search, we select the highest-fitness update rule from the final generation of each evolutionary seed. For GPT-5.2, this yields two distinct algorithms corresponding to the two independent evolutionary runs. Update rules evolved using Claude~4.5~Opus consistently achieved substantially lower fitness and did not produce competitive candidates in the final generation, and are therefore excluded from further analysis.

Each selected algorithm is evaluated on the full suite of ten environments shown in Figure~\ref{fig:env_comparison}, including both the five environments used during evolutionary search and five additional environments not seen during fitness optimization. For each environment, we report the best performance achieved after post-evolution hyperparameter optimization, following the evaluation protocol defined in Section~\ref{sec:method_fitness}. This procedure reflects the maximal performance attainable by a given update rule under a fixed training budget.

The two highest-performing algorithms identified by this process are coined \textbf{Confidence-Guided Forward Policy Distillation (CG-FPD)} and \textbf{Differentiable Forward Confidence-Weighted Planning with Controllability Prior (DF-CWP-CP)}. At a high level, \textbf{CG-FPD} trains a policy by distilling short-horizon plans from a learned latent dynamics model, using planning solely as a supervised teacher signal. In contrast, \textbf{DF-CWP-CP} optimizes the policy via differentiable short-horizon world-model rollouts that incorporate confidence-weighted desirability and controllability objectives. Both algorithms avoid value functions, policy gradients, and Bellman-style updates, and instead rely on planning-derived learning signals. Detailed descriptions of their internal mechanisms are provided in Appendix~\ref{app:best_evolved_algorithms}.

\paragraph{Generalization across environments.}
Evolutionary fitness is computed on only five training environments, which may lead to overfitting. To assess generalization, we evaluate the two selected algorithms, \textbf{CG-FPD} and \textbf{DF-CWP-CP}, on ten environments (Figure~\ref{fig:env_comparison}), including the five training and five unseen ones. All evaluations follow the same protocol as during evolution, and results report the best performance after post-evolution hyperparameter tuning. We compare against PPO, A2C, DQN, and SAC baselines using an identical $256 \times 256$ MLP policy architecture (Appendix~\ref{app: fair comparison}). Table~\ref{tab:results} shows the maximum evaluation return per environment.


\begin{table*}[t]
\centering
\resizebox{\textwidth}{!}{
\begin{tabular}{lcccccc}
\toprule
\textbf{Environment} &
\textbf{PPO} &
\textbf{A2C} &
\textbf{DQN} &
\textbf{SAC} &
\textbf{CG-FPD} &
\textbf{DF-CWP-CP} \\
\midrule

CartPole
& $\mathbf{500.0 \pm 0.0}$
& $\mathbf{500.0 \pm 0.0}$
& $\mathbf{500.0 \pm 0.0}$
& --
& $\mathbf{500.0 \pm 0.0}$
& $\mathbf{500.0 \pm 0.0}$ \\

LunarLander
& $246.60 \pm 30.9$
& $246.60 \pm 13.4$
& $250.10 \pm 4.10$
& --
& $241.20 \pm 11.0$
& $\mathbf{260.60 \pm 19.12}$ \\

MountainCar
& $-128.12 \pm 44.64$
& $-134.20 \pm 8.70$
& $-147.80 \pm 40.40$
& --
& $\mathbf{-105.80 \pm 10.51}$
& $-108.67 \pm 7.34$ \\

Acrobot
& $-63.5 \pm 0.8$
& $-213.6 \pm 202.5$
& $\mathbf{-61.90 \pm 0.10}$
& --
& $-90.6 \pm 5.86$
& $-78.67 \pm 8.72$ \\

HalfCheetah
& $1579.13 \pm 643.80$
& $795.91 \pm 147.70$
& --
& $\mathbf{4988.5 \pm 2667.70}$
& $2407.88 \pm 311.90$
& $2103.80 \pm 246.70$ \\

Reacher
& $-3.15 \pm 0.14$
& $-6.48 \pm 0.57$
& --
& $\mathbf{-2.05 \pm 0.19}$
& $-2.67 \pm 0.15$
& $-5.43 \pm 0.44$ \\

Swimmer
& $95.0 \pm 29.20$
& $49.08 \pm 1.07$
& --
& $87.84 \pm 23.50$
& $\mathbf{247.54 \pm 35.53}$
& $219.82 \pm 32.33$ \\

Inverted Pendulum
& $\mathbf{1000.0 \pm 0.0}$
& $\mathbf{1000.0 \pm 0.0}$
& --
& $\mathbf{1000.0 \pm 0.0}$
& $\mathbf{1000.0 \pm 0.0}$
& $\mathbf{1000.0 \pm 0.0}$ \\

Walker2d
& $3163.20 \pm 397.20$
& $801.50 \pm 291.10$
& --
& $\mathbf{4595.30 \pm 252.20}$
& $1603.80 \pm 146.70$
& $1297.63 \pm 62.77$ \\

Pusher
& $\mathbf{-25.50 \pm 0.32}$
& $-32.41 \pm 0.81$
& --
& $\mathbf{-25.50 \pm 0.32}$
& $-27.23 \pm 0.86$
& $-39.88 \pm 0.92$ \\

\bottomrule
\end{tabular}
}
\caption{Mean $\pm$ standard deviation of the maximum evaluation return per environment. Each algorithm is trained with five seeds; for each seed, the checkpoint with the highest evaluation return is evaluated over 100 episodes, and results report the mean and standard deviation across seeds. Dashes (--) indicate inapplicability (e.g., SAC for discrete environments, DQN for continuous control). PPO and A2C support both action types and are evaluated alongside the evolved methods (\textbf{CG-FPD} and DF-CWP-CP) on all compatible environments.}
\label{tab:results}
\end{table*}

\begin{figure*}[t]
    \centering
    \setlength{\tabcolsep}{2pt}
    \renewcommand{\arraystretch}{1.1}

    \begin{tabular}{ccccc}
        \multicolumn{5}{c}{\textbf{CG-FPD}} \\
        CartPole \hspace{52pt} &
        LunarLander \hspace{52pt} &
        MountainCar \hspace{52pt} &
        Acrobot \hspace{52pt} &
        HalfCheetah
        \hspace{-18pt}
    \end{tabular}

    \vspace{2pt}

    \begin{tabular}{ccccc}
        \includegraphics[width=0.19\textwidth]{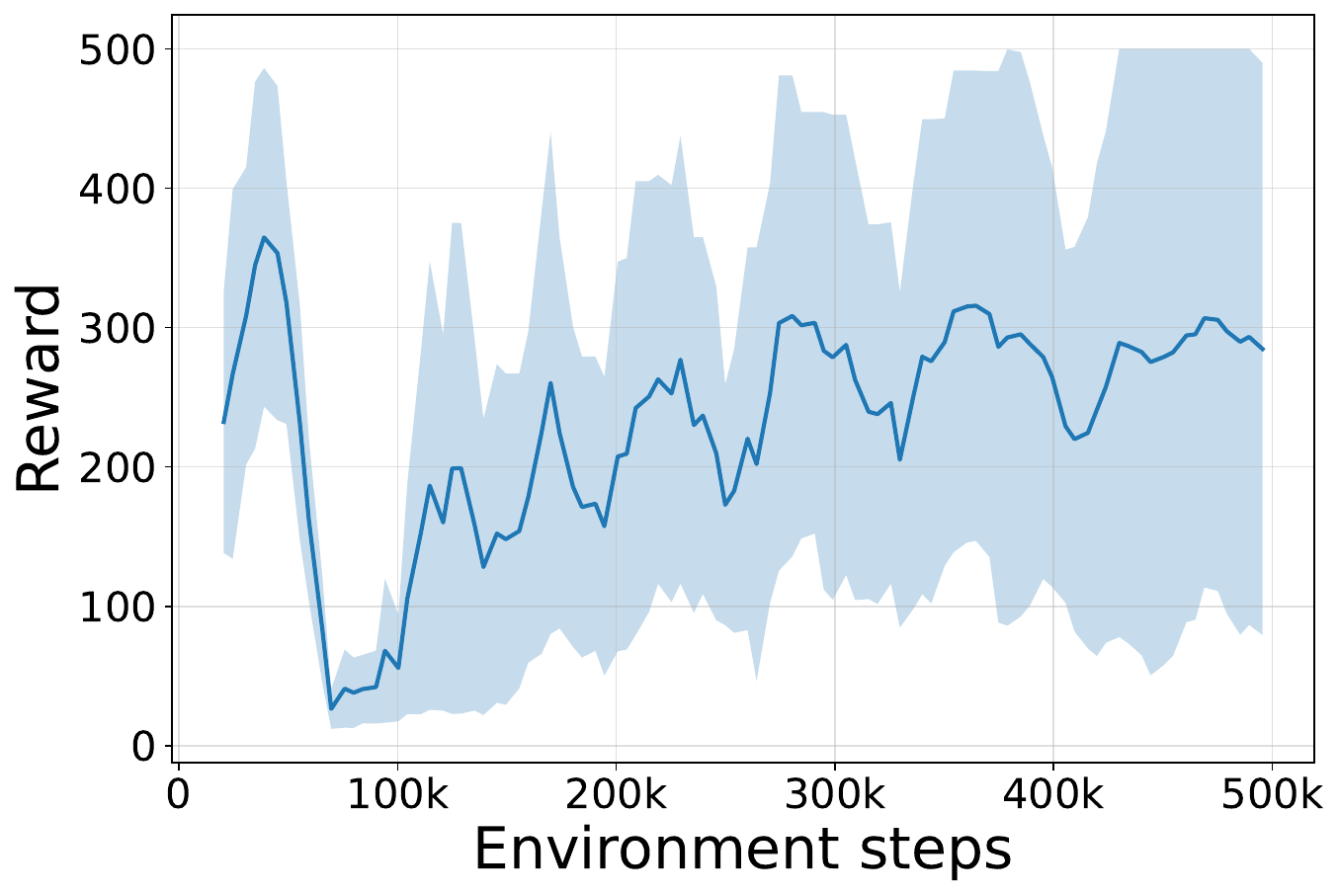} &
        \includegraphics[width=0.19\textwidth]{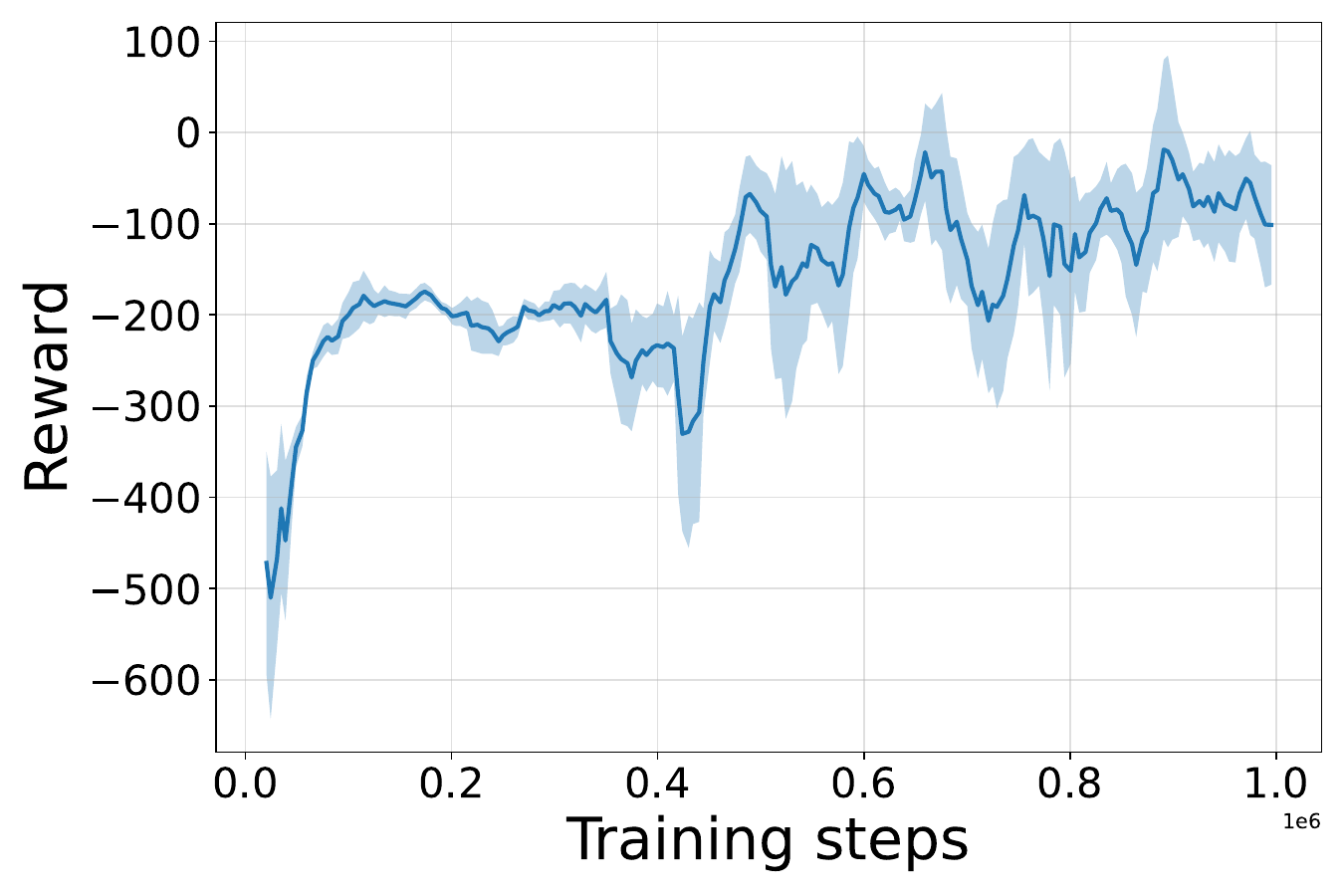} &
        \includegraphics[width=0.19\textwidth]{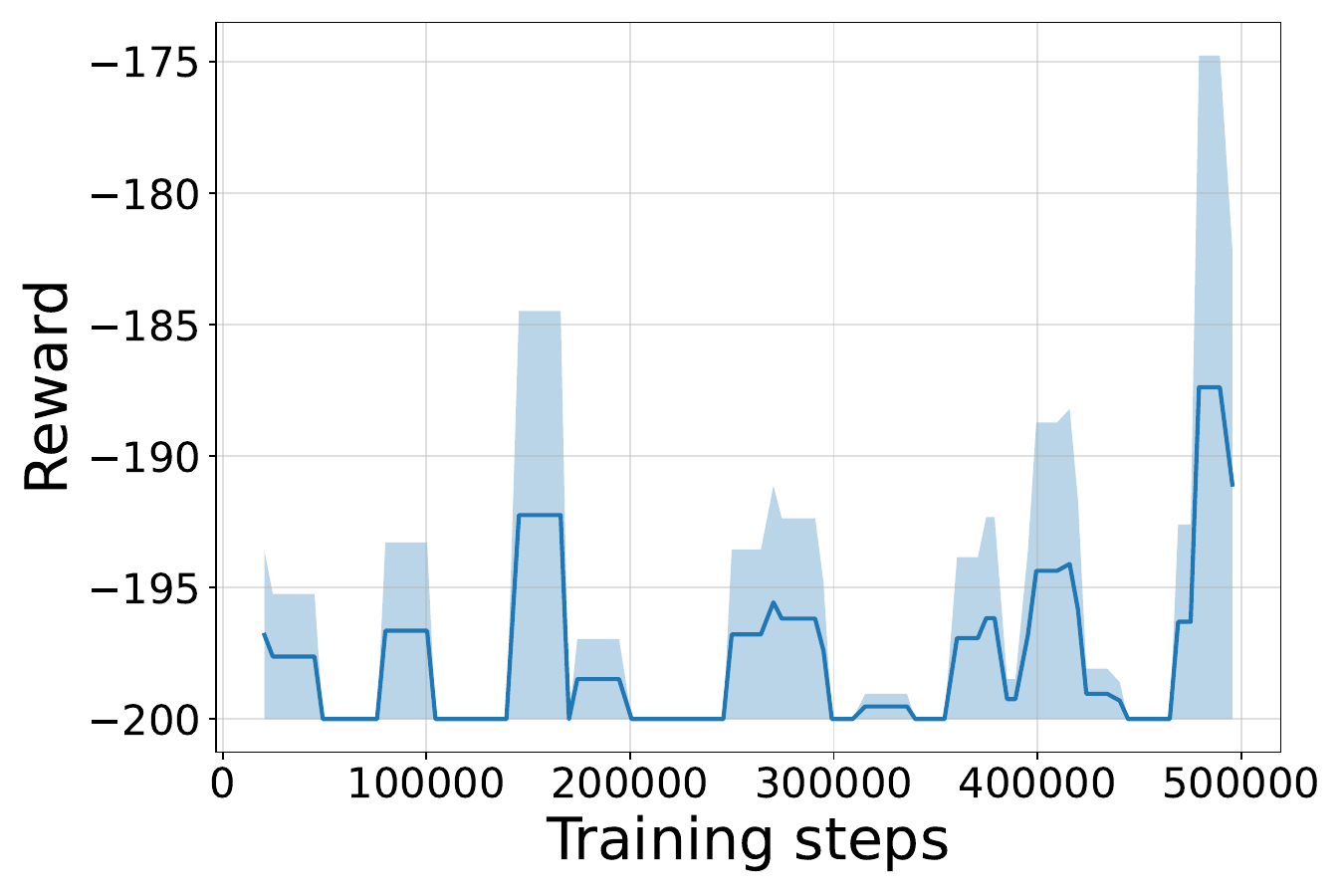} &
        \includegraphics[width=0.19\textwidth]{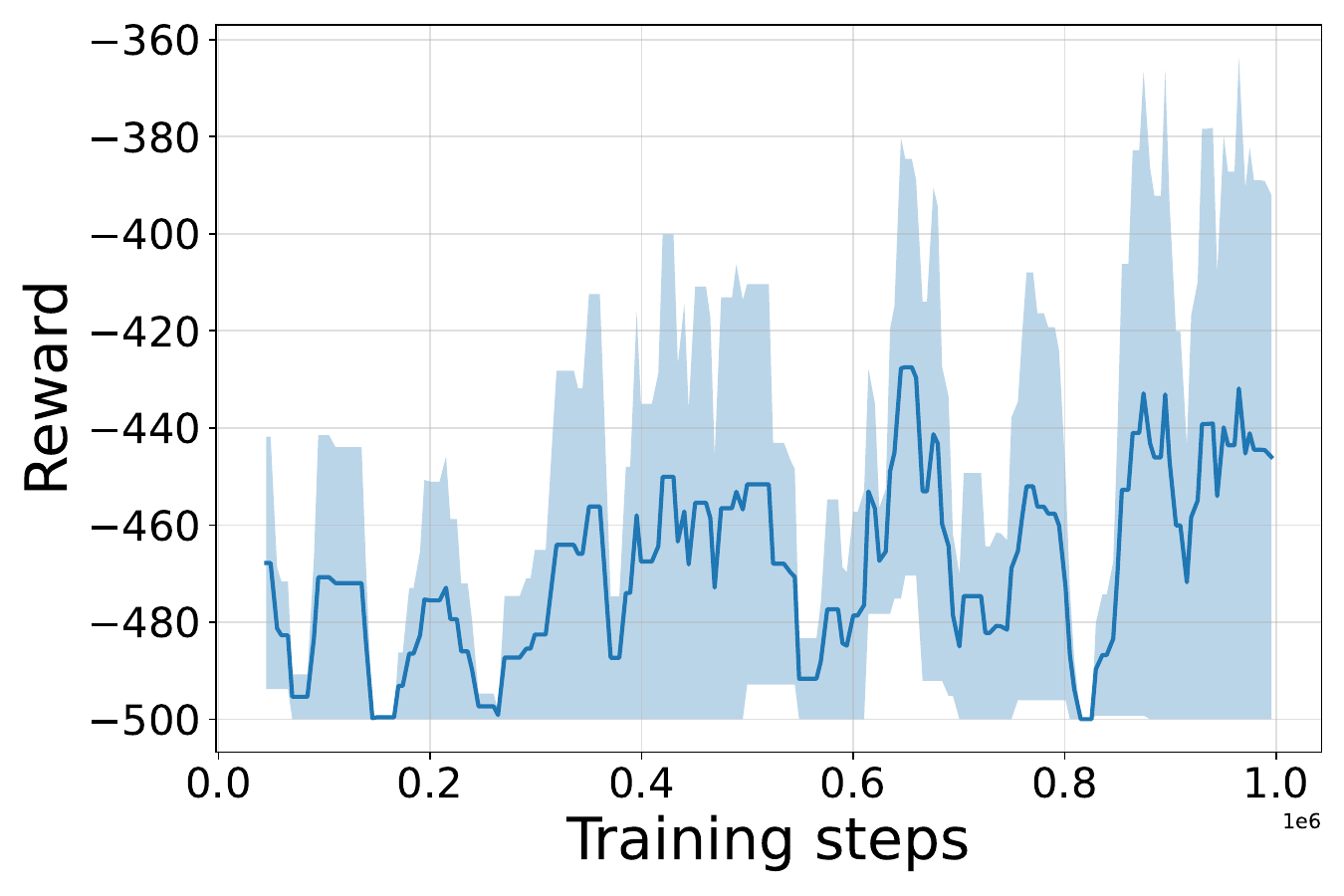} &
        \includegraphics[width=0.19\textwidth]{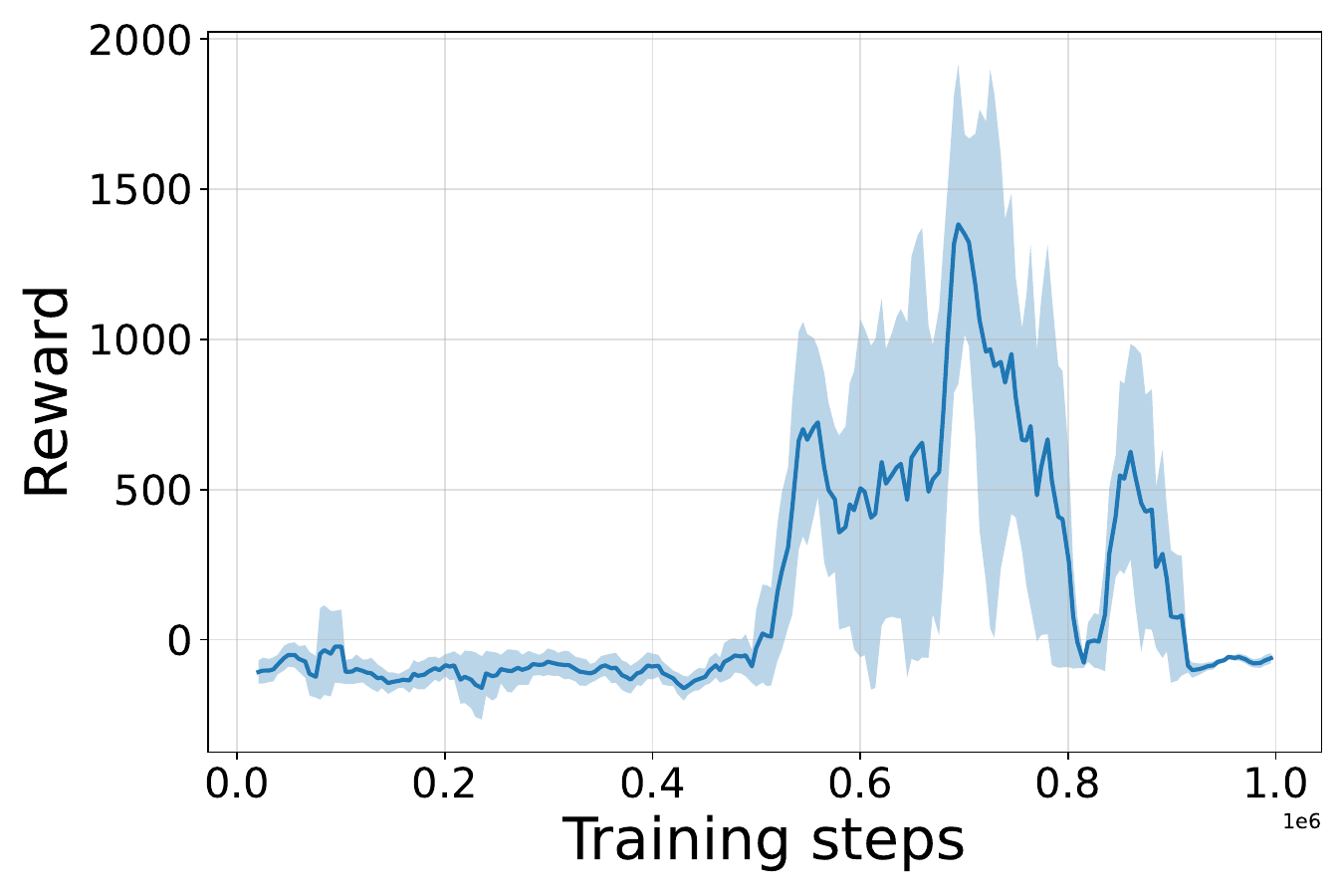}
    \end{tabular}

    \vspace{3pt}

    \begin{tabular}{ccccc}
        Walker2d \hspace{48pt} &
        InvertedPendulum \hspace{42pt} &
        Reacher \hspace{64pt} &
        Swimmer \hspace{64pt} &
        Pusher
    \end{tabular}

    \vspace{2pt}

    \begin{tabular}{ccccc}
        \includegraphics[width=0.19\textwidth]{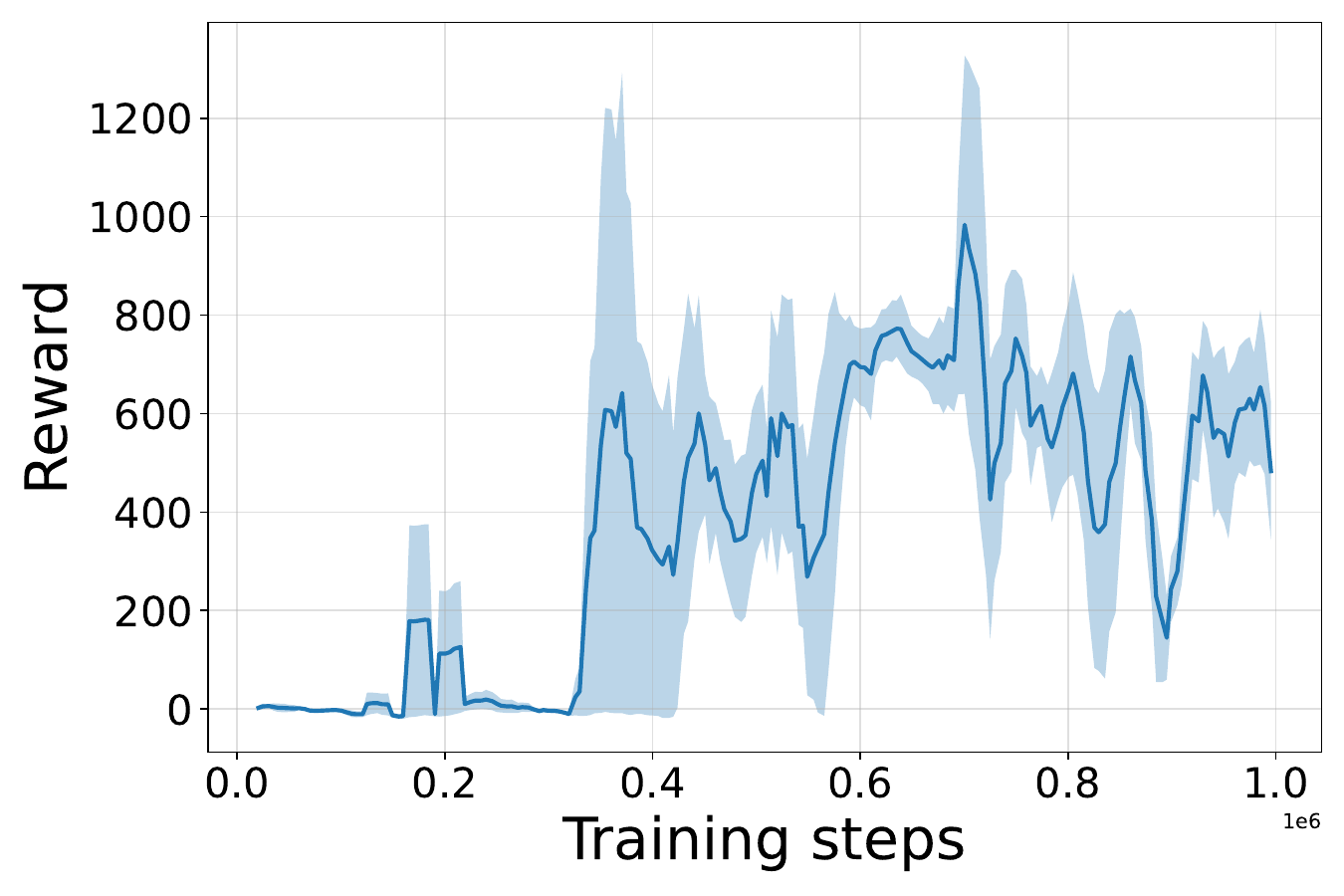} &
        \includegraphics[width=0.19\textwidth]{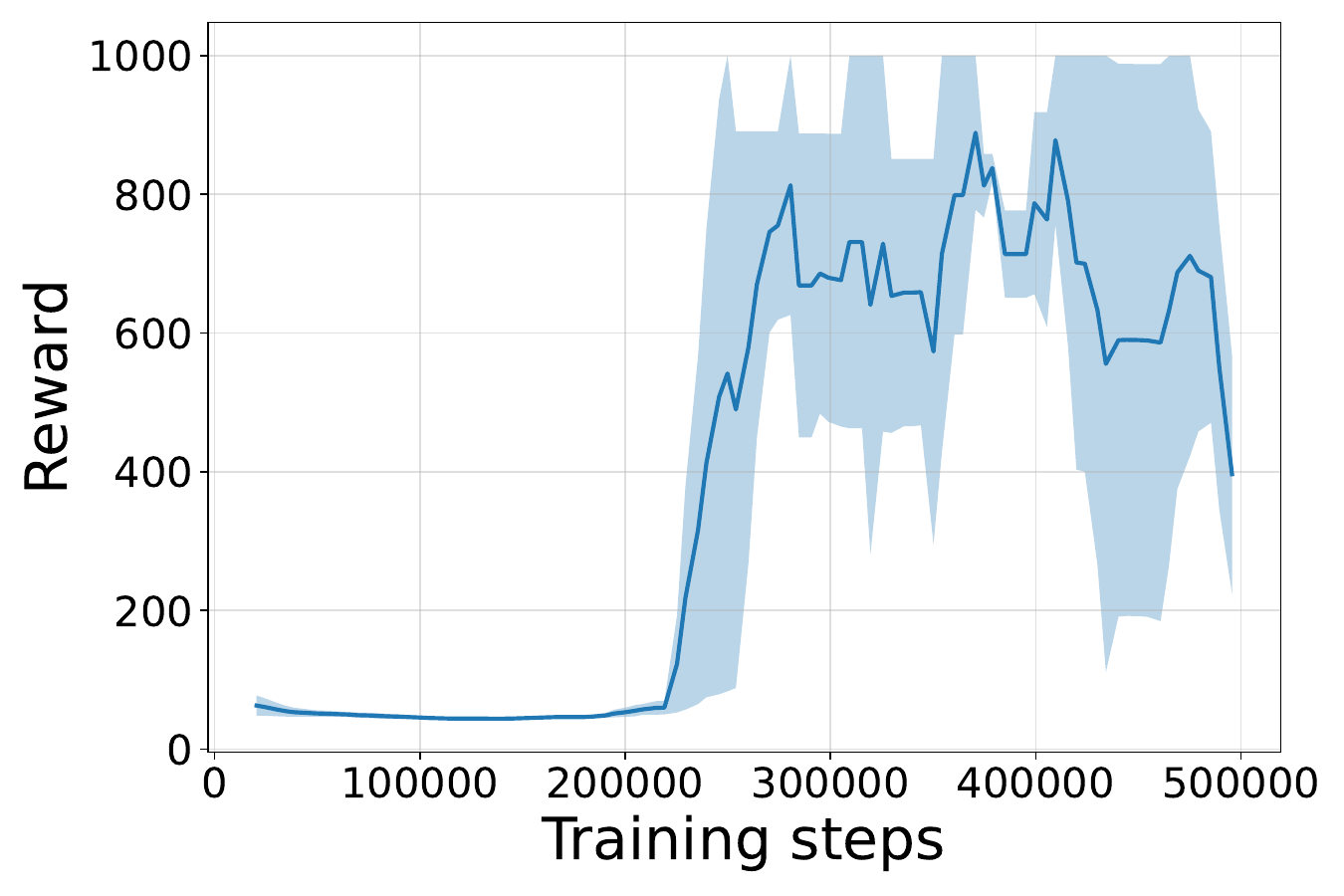} &
        \includegraphics[width=0.19\textwidth]{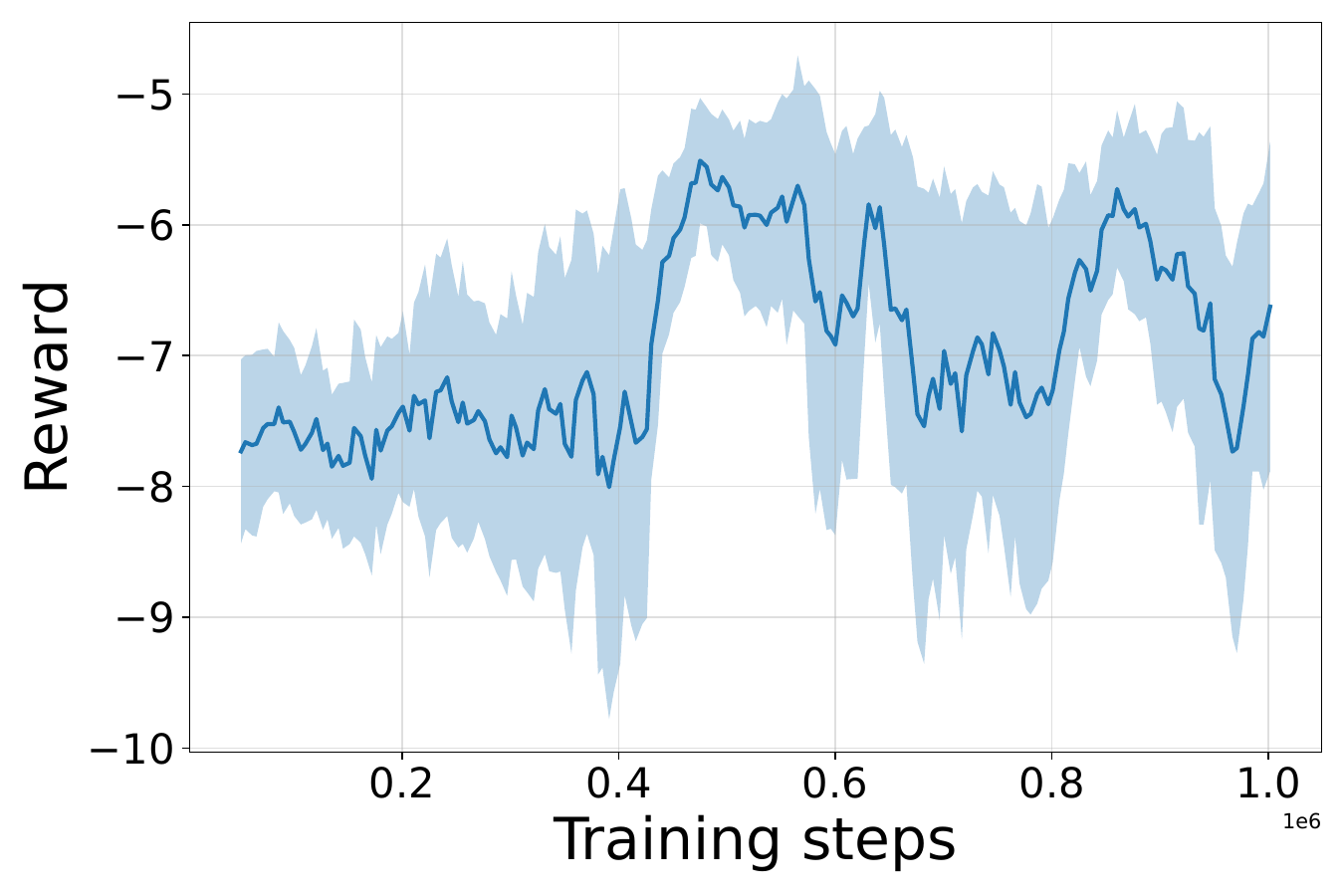} &
        \includegraphics[width=0.19\textwidth]{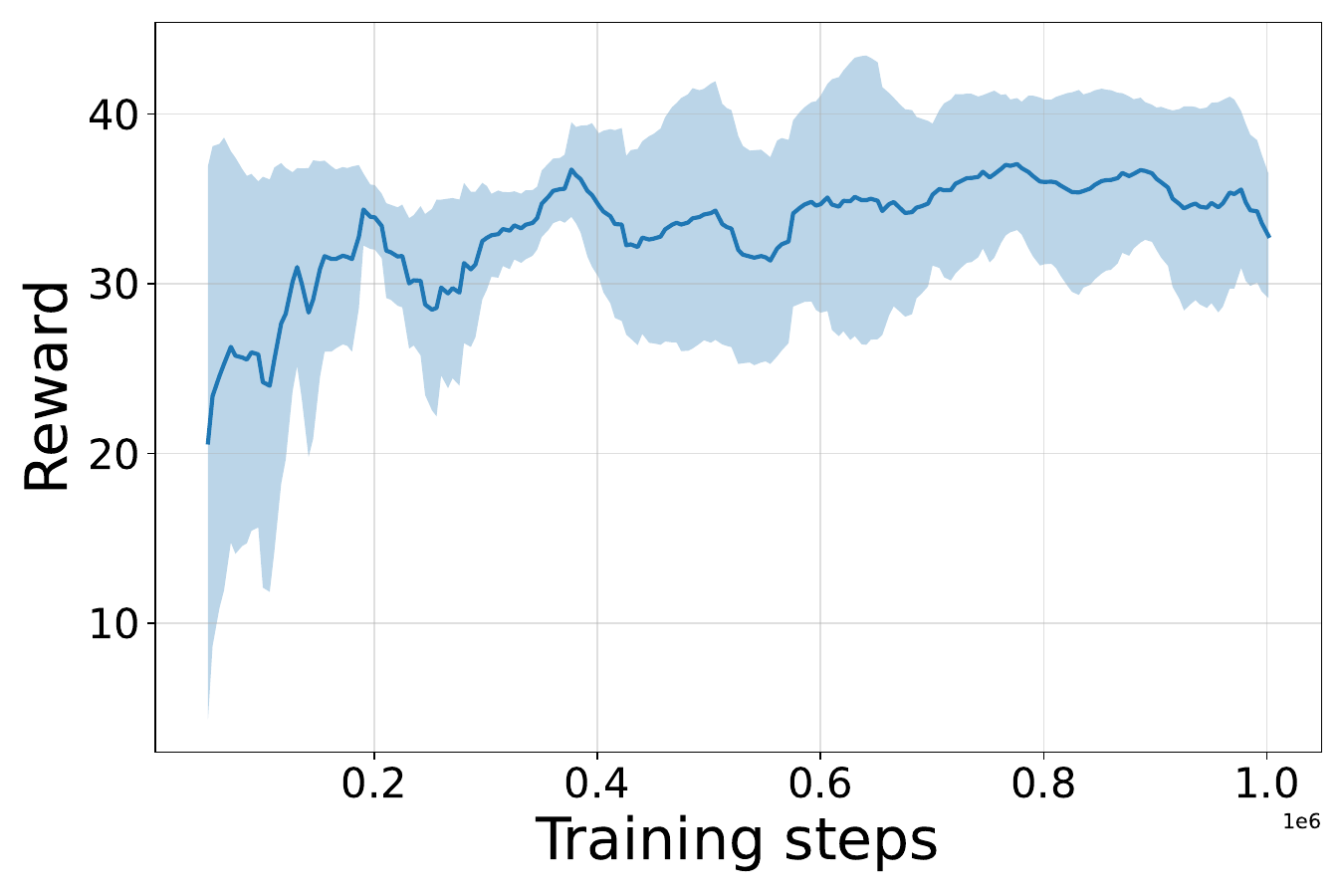} &
        \includegraphics[width=0.19\textwidth]{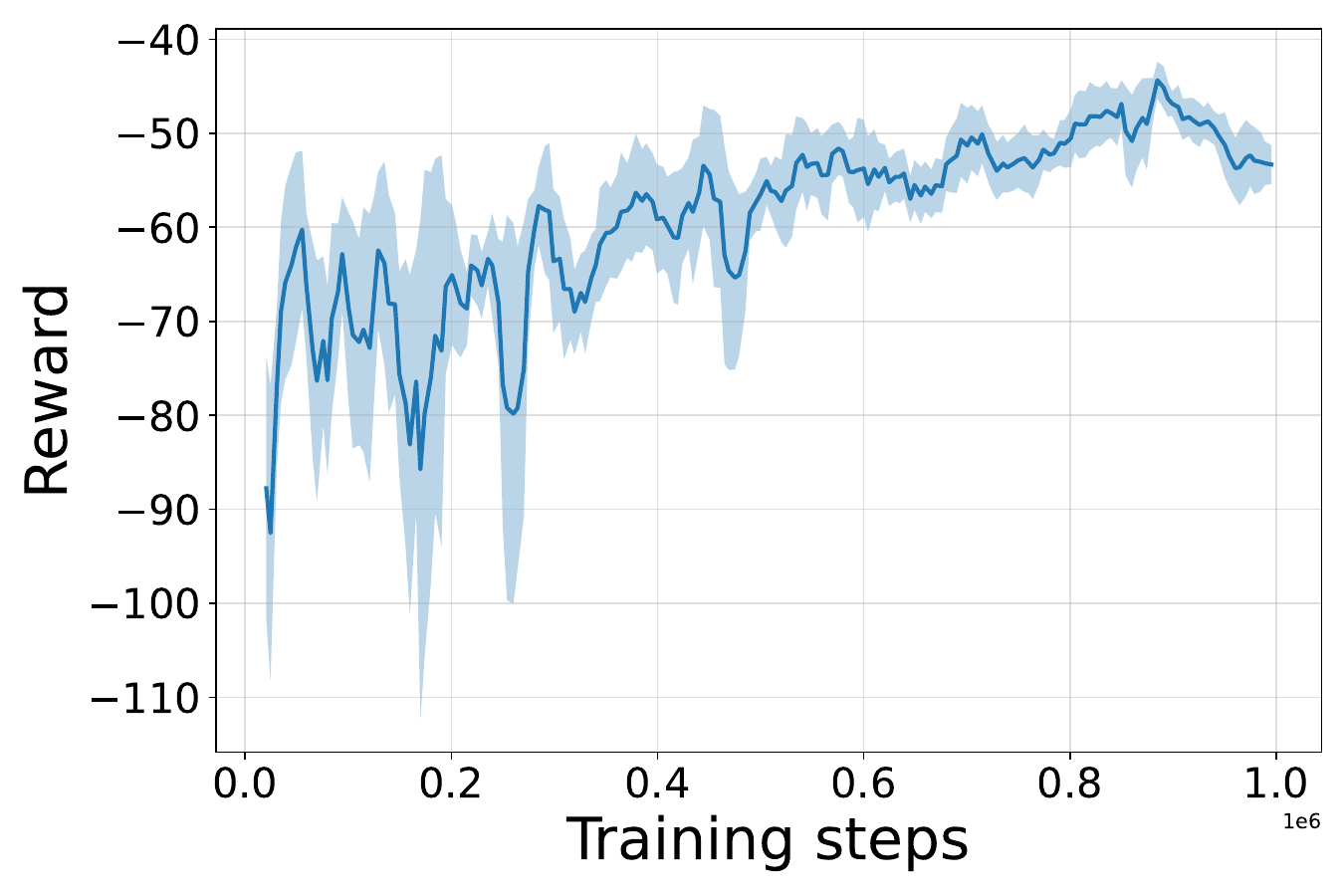}
    \end{tabular}

    \vspace{10pt}

    \begin{tabular}{ccccc}
        \multicolumn{5}{c}{\textbf{DF-CWP-CP}} \\
        CartPole \hspace{52pt} &
        LunarLander \hspace{52pt} &
        MountainCar \hspace{52pt} &
        Acrobot \hspace{52pt} &
        HalfCheetah
        \hspace{-18pt}
    \end{tabular}

    \vspace{2pt}

    \begin{tabular}{ccccc}
        \includegraphics[width=0.19\textwidth]{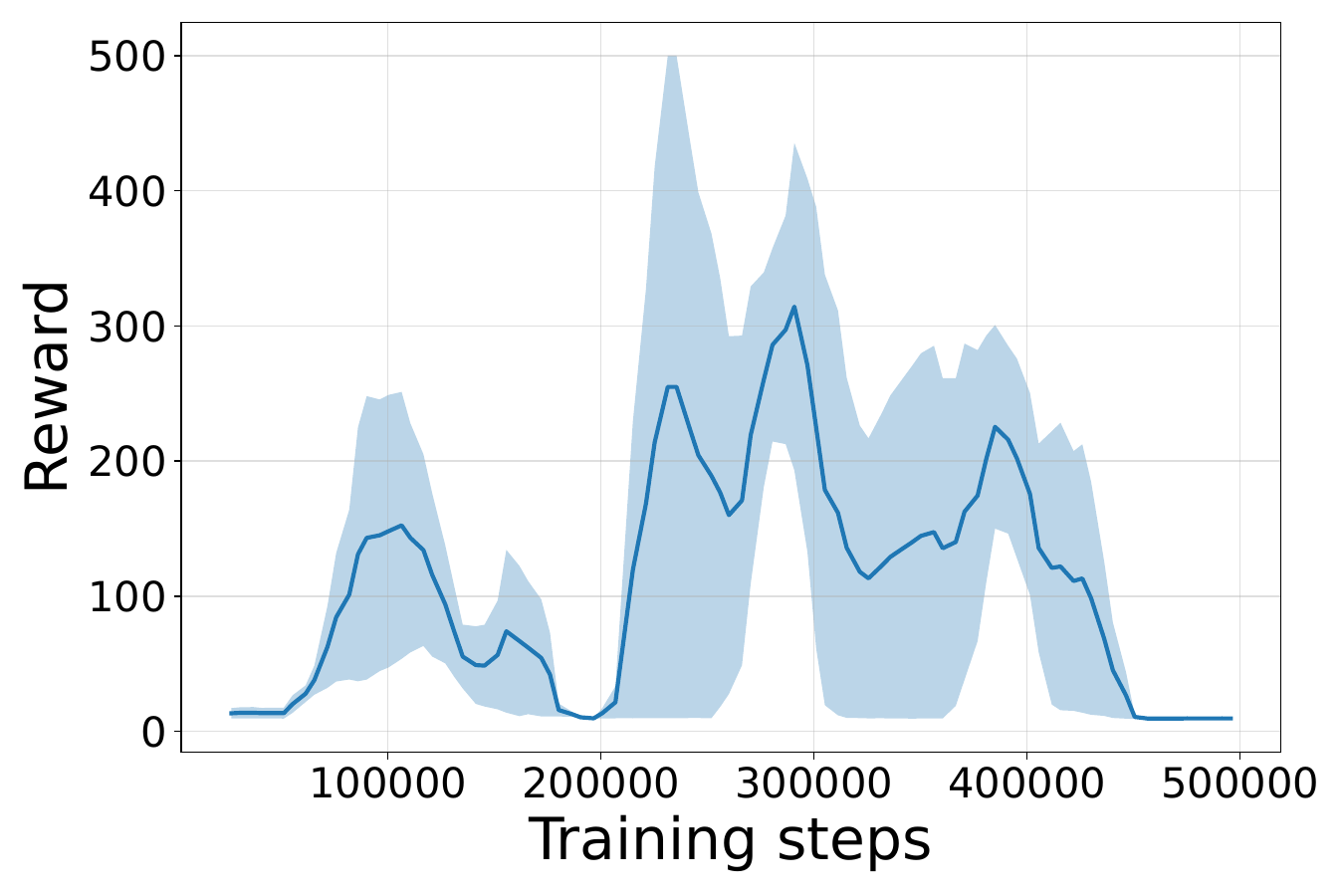} &
        \includegraphics[width=0.19\textwidth]{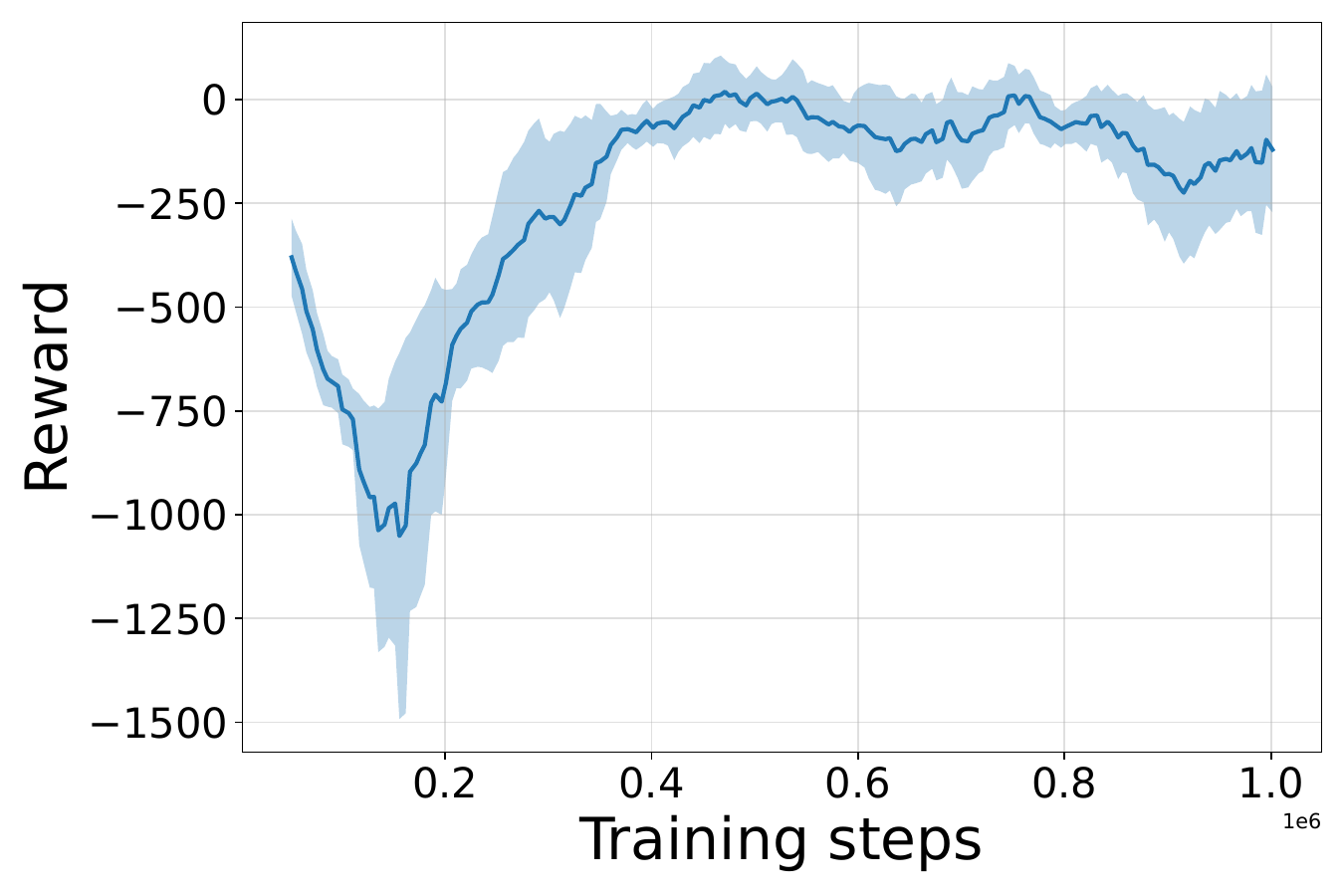} &
        \includegraphics[width=0.19\textwidth]{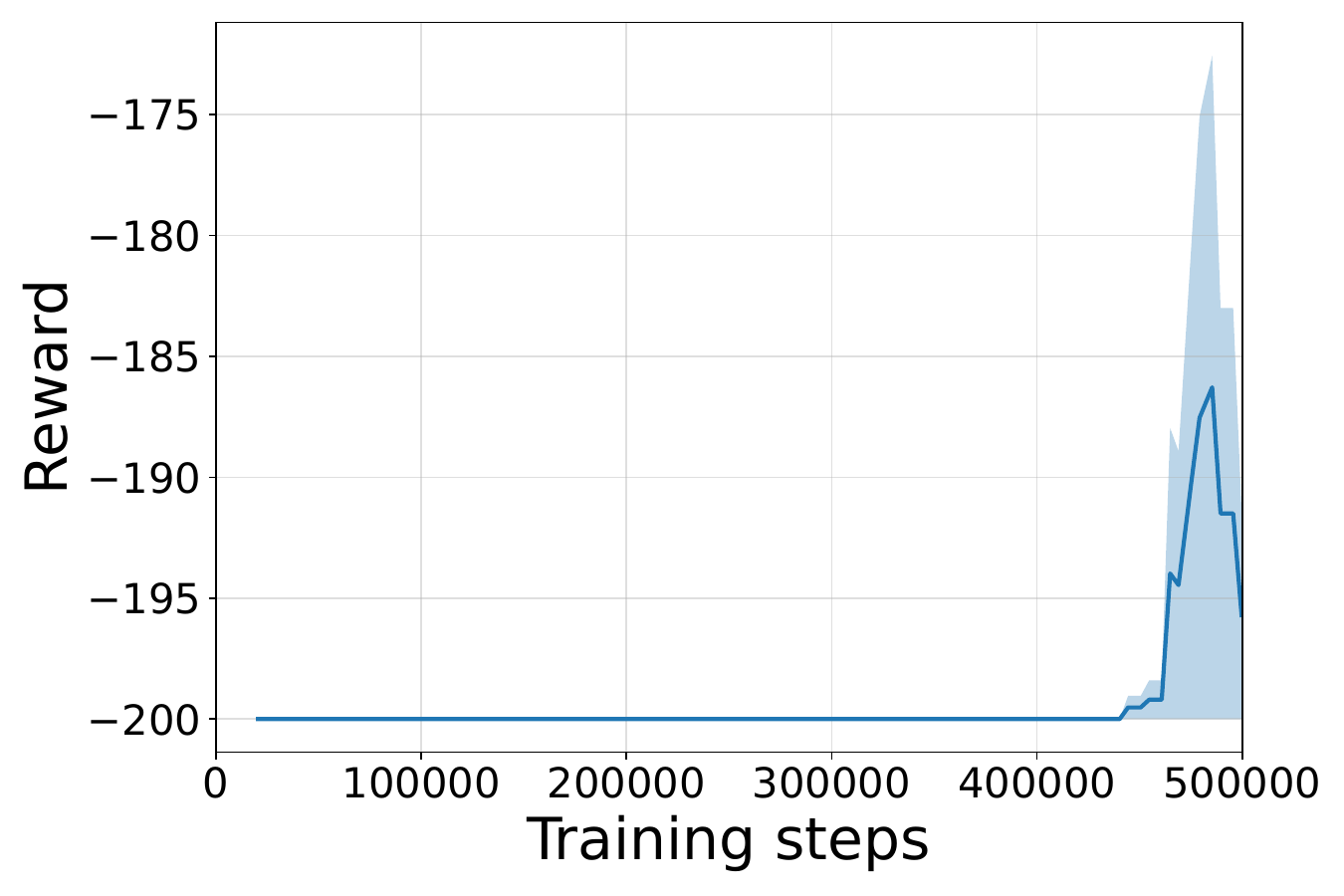} &
        \includegraphics[width=0.19\textwidth]{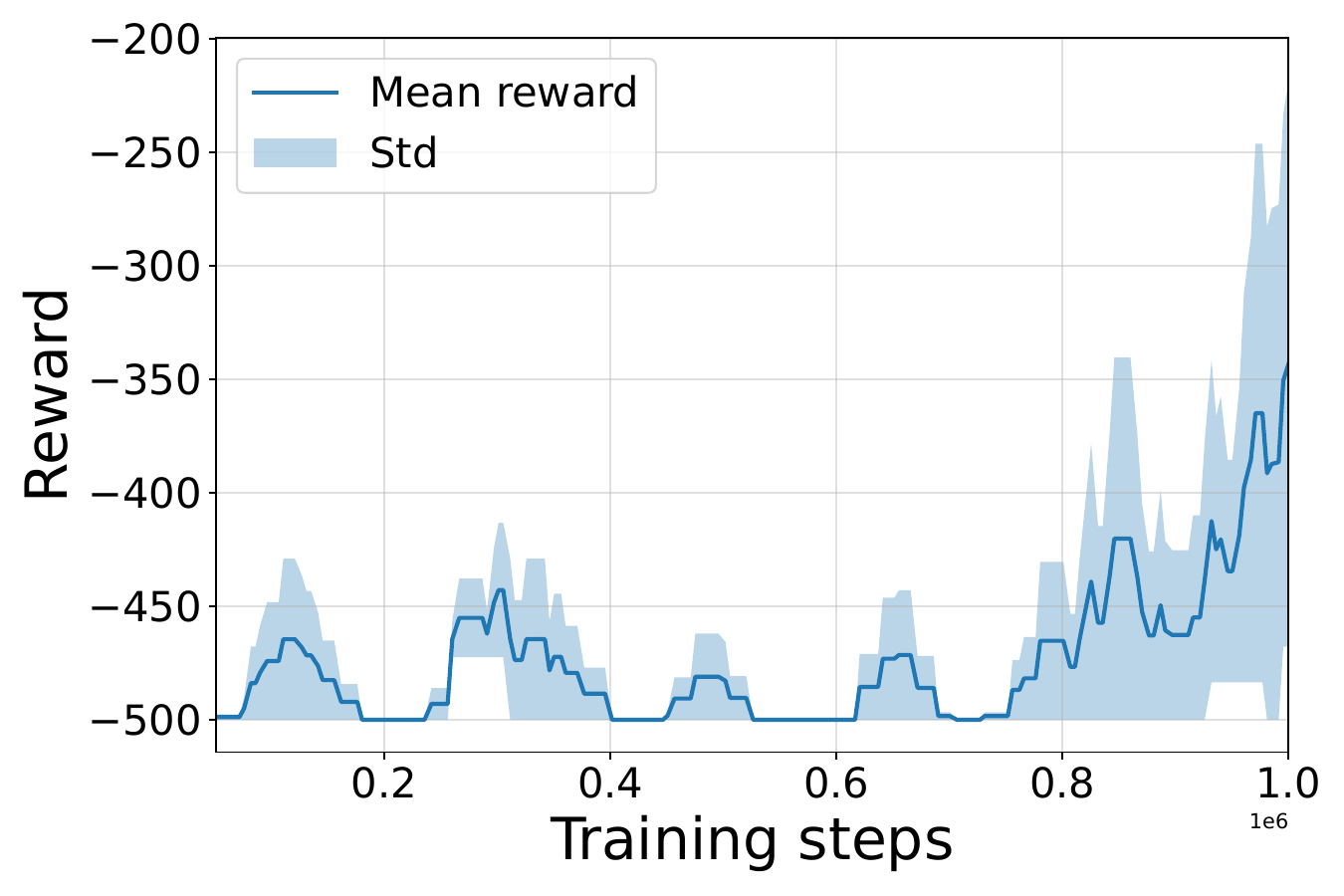} &
        \includegraphics[width=0.19\textwidth]{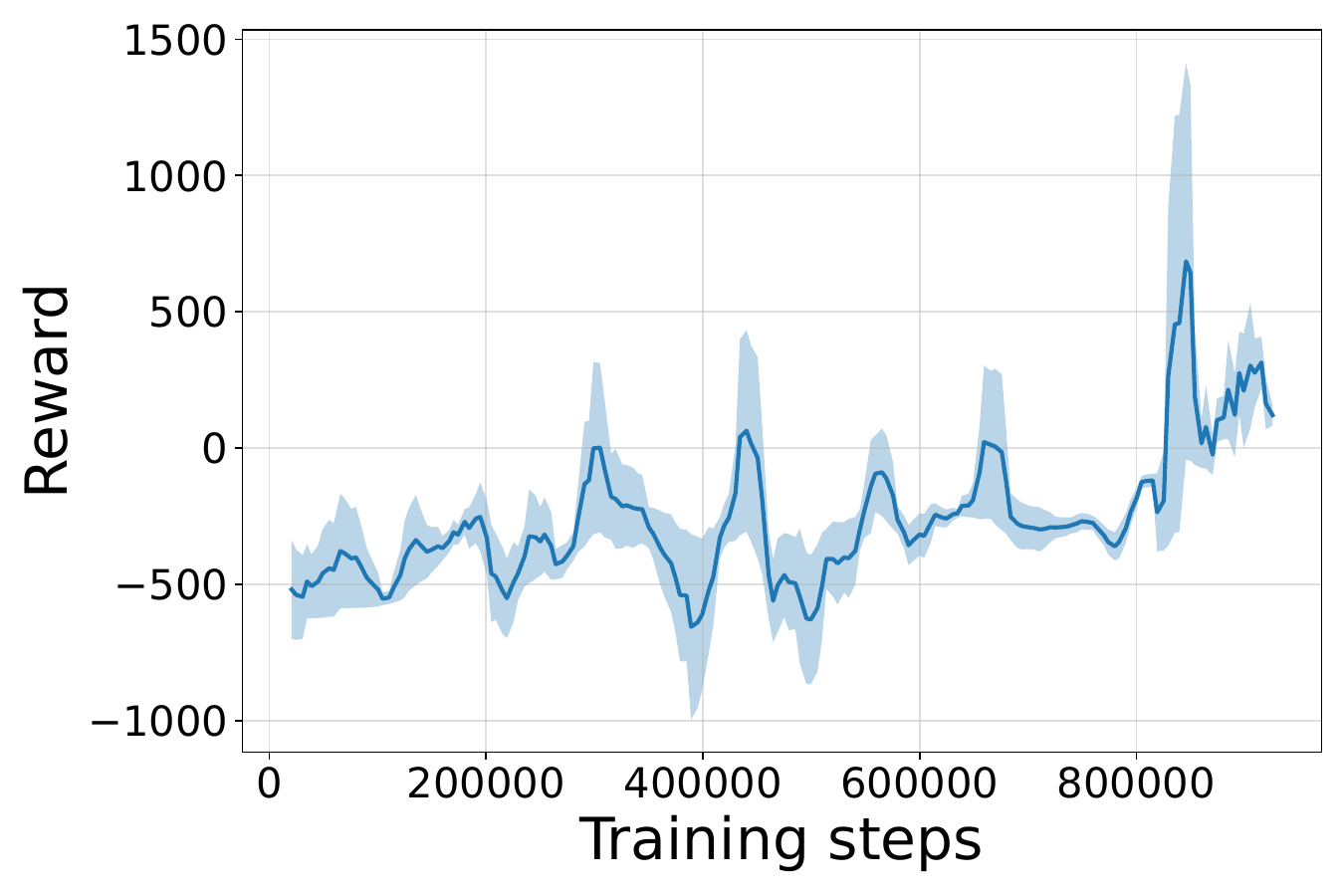}
    \end{tabular}

    \vspace{3pt}

    \begin{tabular}{ccccc}
        Walker2d \hspace{48pt} &
        InvertedPendulum \hspace{42pt} &
        Reacher \hspace{64pt} &
        Swimmer \hspace{64pt} &
        Pusher
    \end{tabular}

    \vspace{2pt}

    \begin{tabular}{ccccc}
        \includegraphics[width=0.19\textwidth]{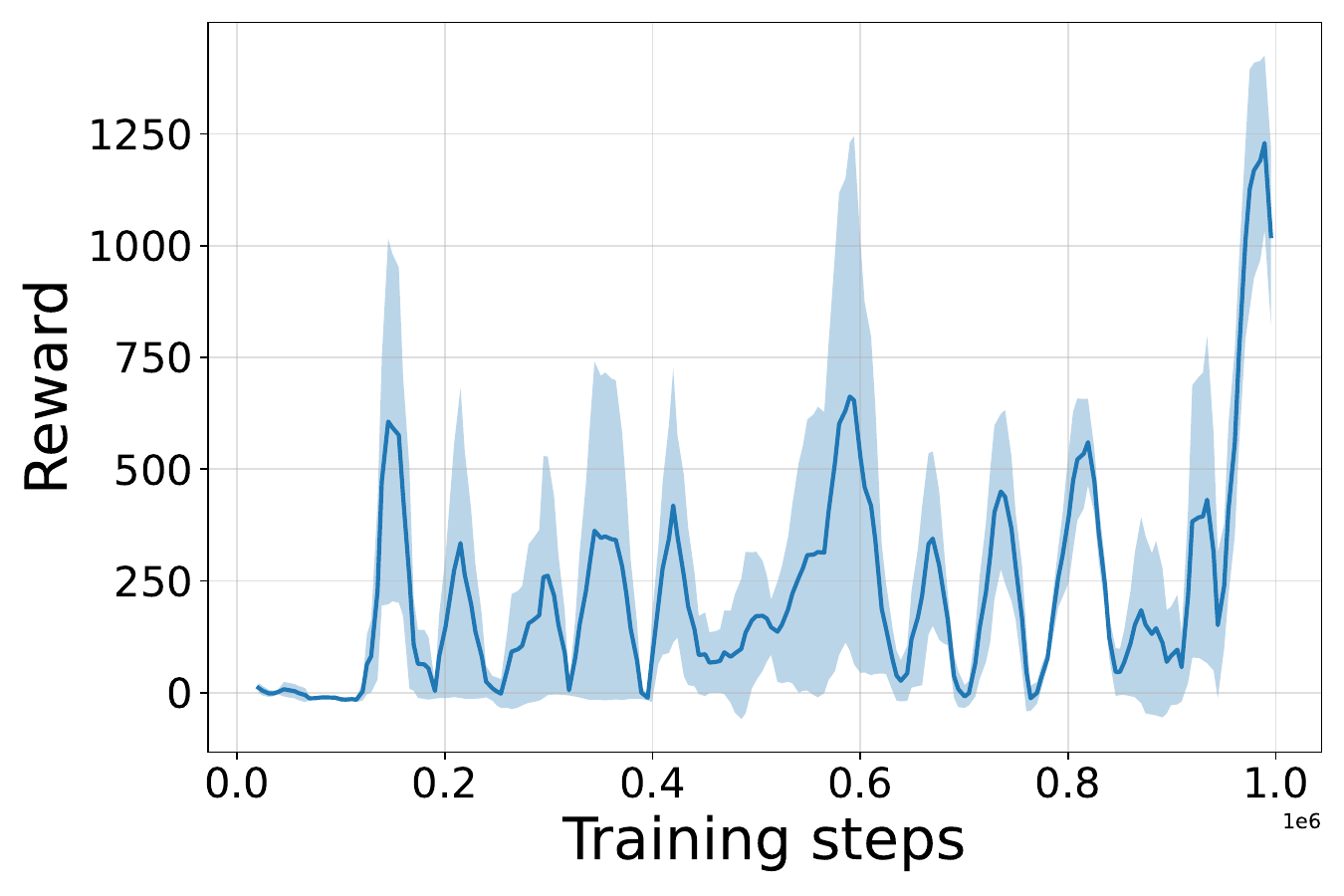} &
        \includegraphics[width=0.19\textwidth]{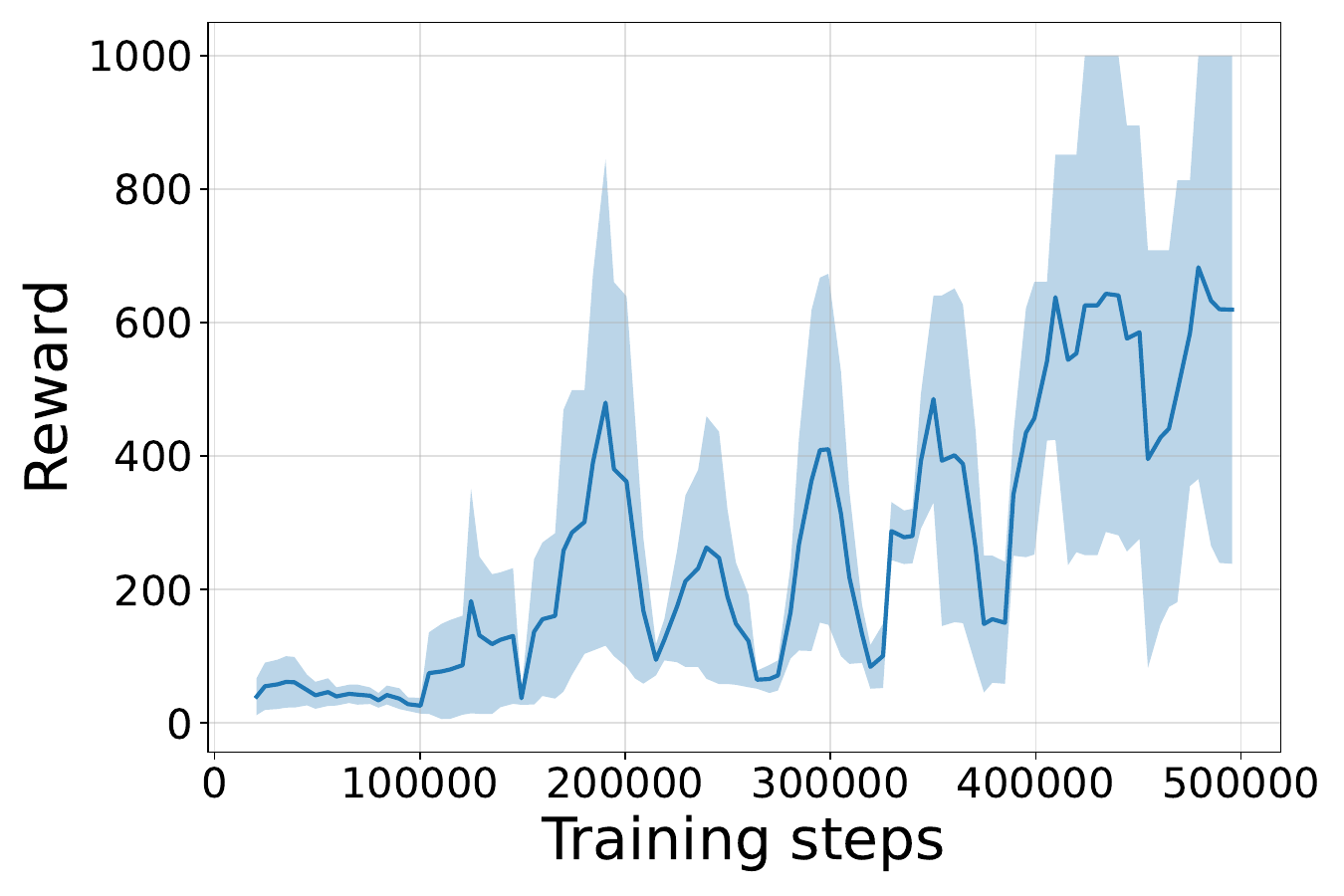} &
        \includegraphics[width=0.19\textwidth]{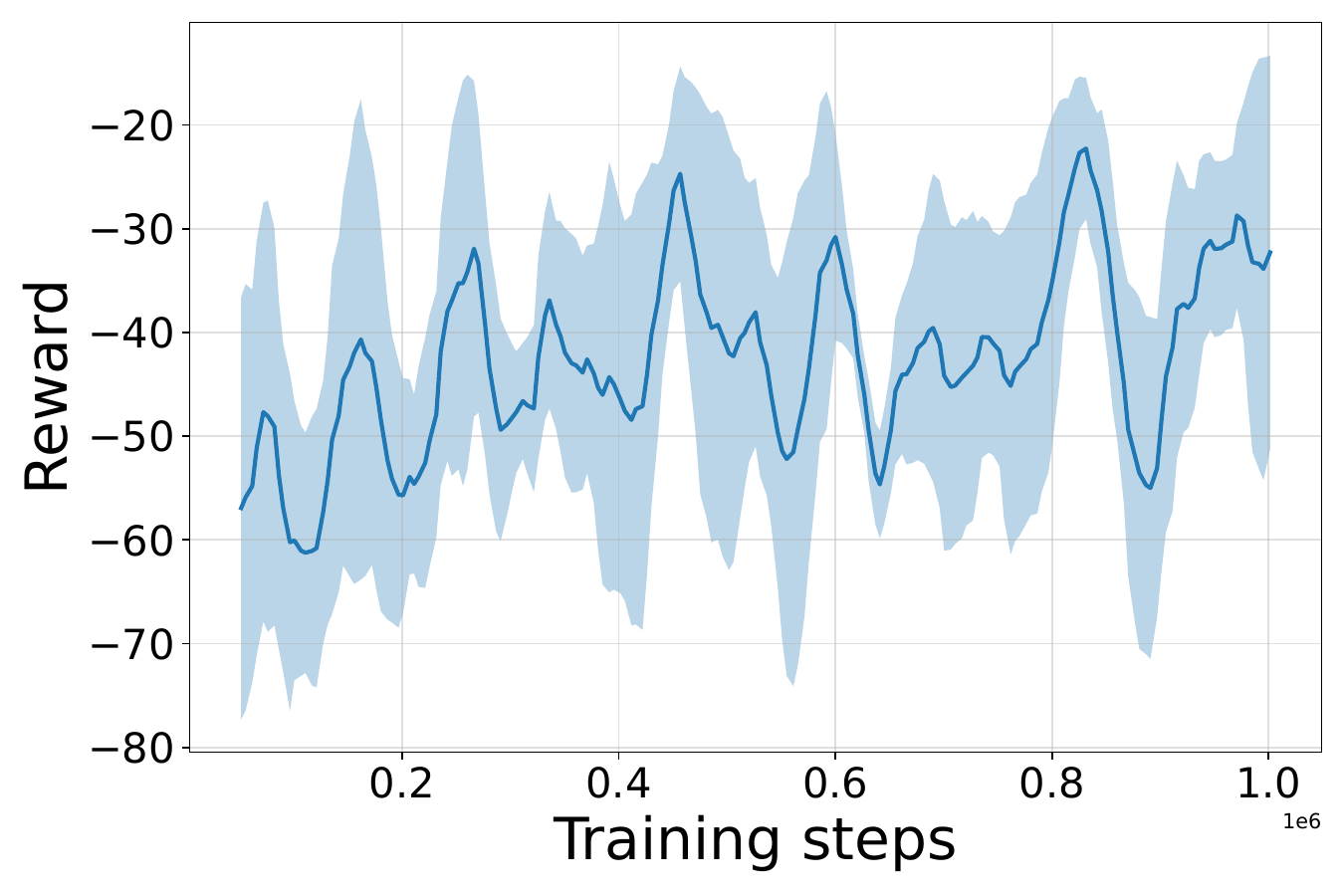} &
        \includegraphics[width=0.19\textwidth]{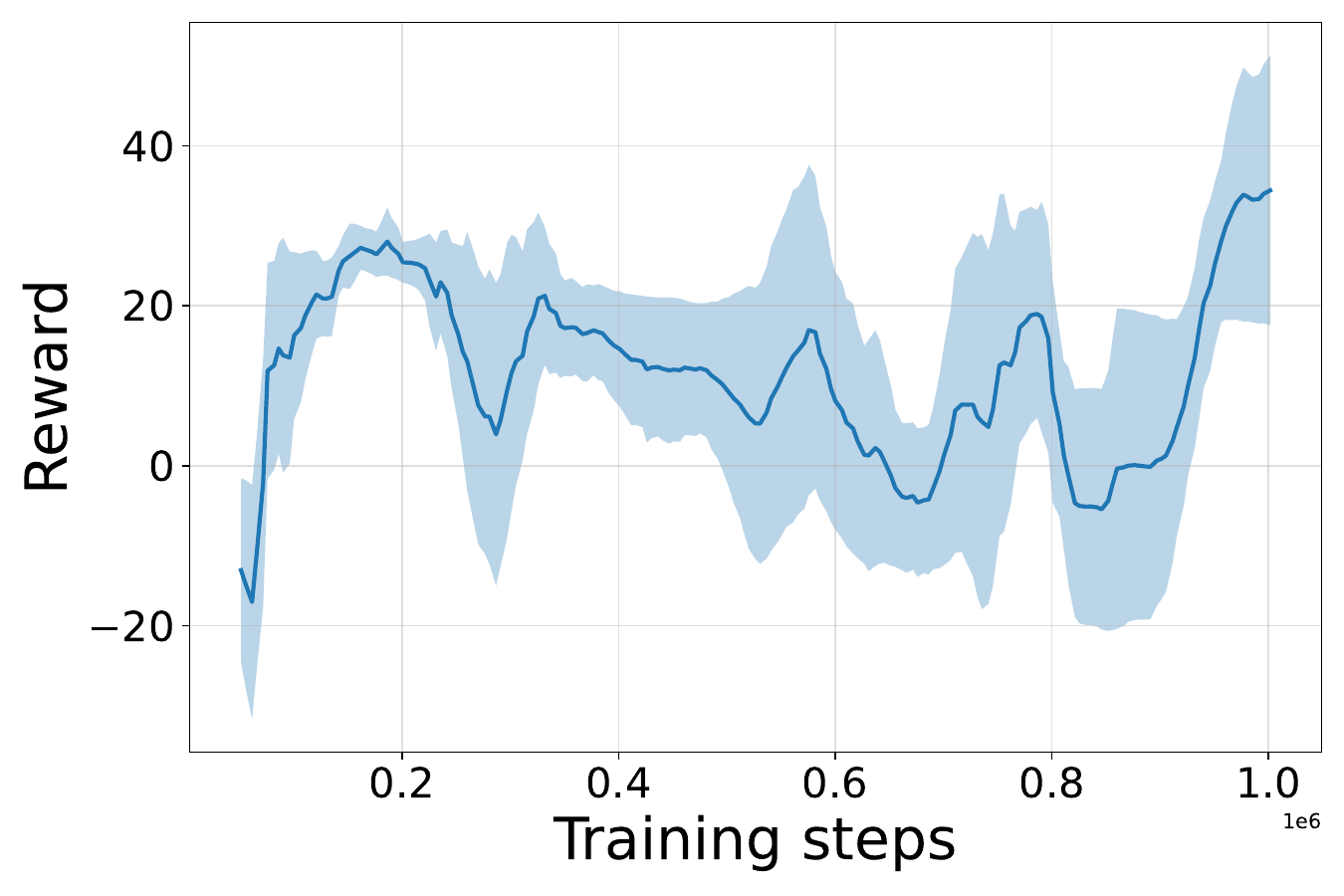} &
        \includegraphics[width=0.19\textwidth]{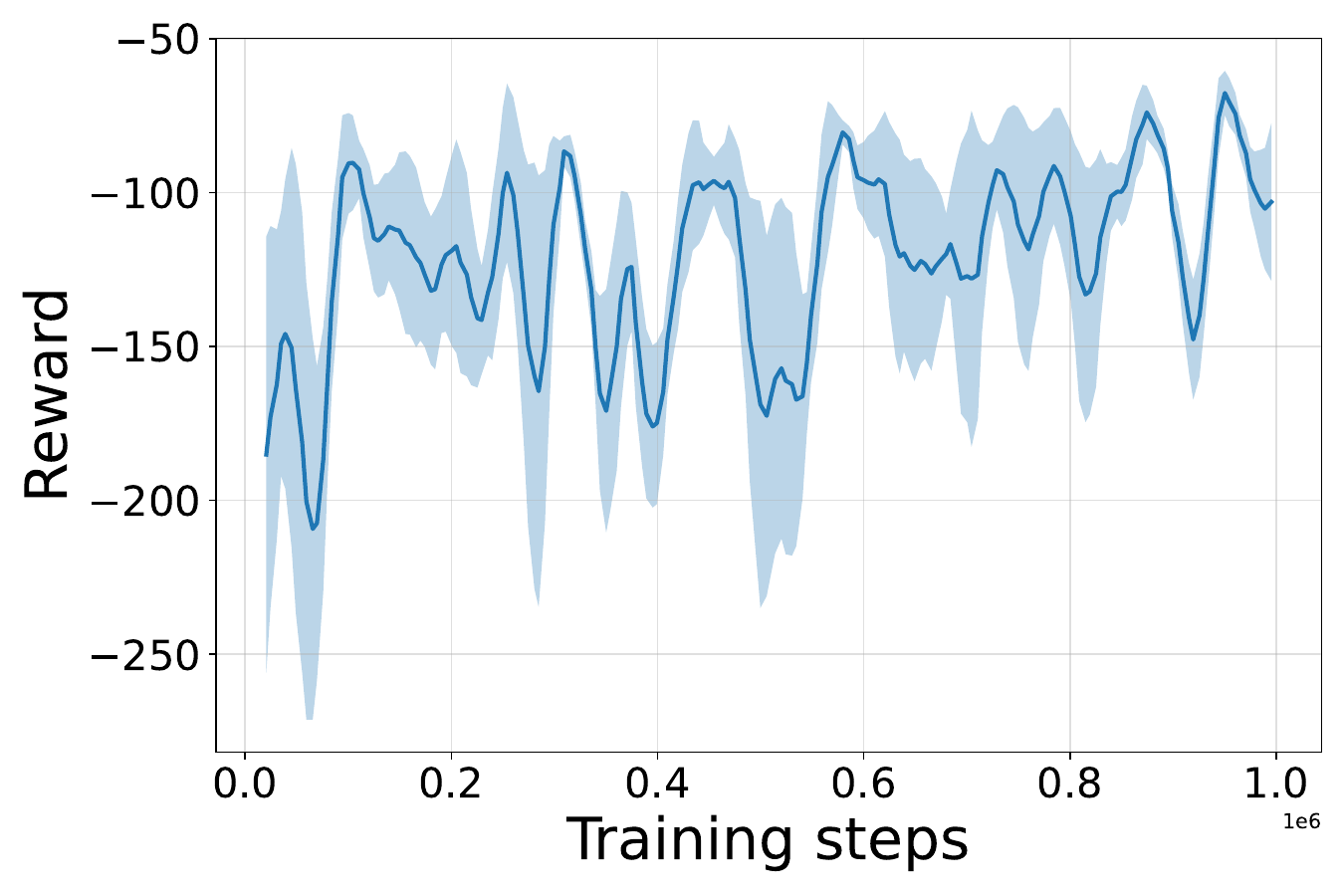}
    \end{tabular}
\caption{Seed-averaged evaluation learning curves for the two evolved algorithms across ten environments. Top: \textbf{CG-FPD}. Bottom: DF-CWP-CP. Curves show evaluation return smoothed with a moving average; shaded regions indicate one standard deviation across seeds. CartPole, InvertedPendulum, and MountainCar are trained for 500k steps, while all other environments are trained for 1M steps. Peak evaluation returns observed at the per-seed level are attenuated in the seed-averaged smoothed curves.}
    \label{fig:plots all}
\end{figure*}

\paragraph{Training stability.}
Beyond final performance, we analyze learning dynamics to assess stability. Figure~\ref{fig:plots all} shows seed-averaged evaluation curves for the two best evolved algorithms. In most environments, both reach high performance within the training budget, but learning is often non-monotonic and not consistently maintained at the end. Despite this, the best checkpoints (Table~\ref{tab:results}) correspond to policies that solve or perform competitively on the tasks.

\subsection{Ablation Studies}

\paragraph{Sensitivity to the Levenshtein weight $\alpha$.}

\begin{figure}[H]
    \centering
    \includegraphics[width=0.9\linewidth]{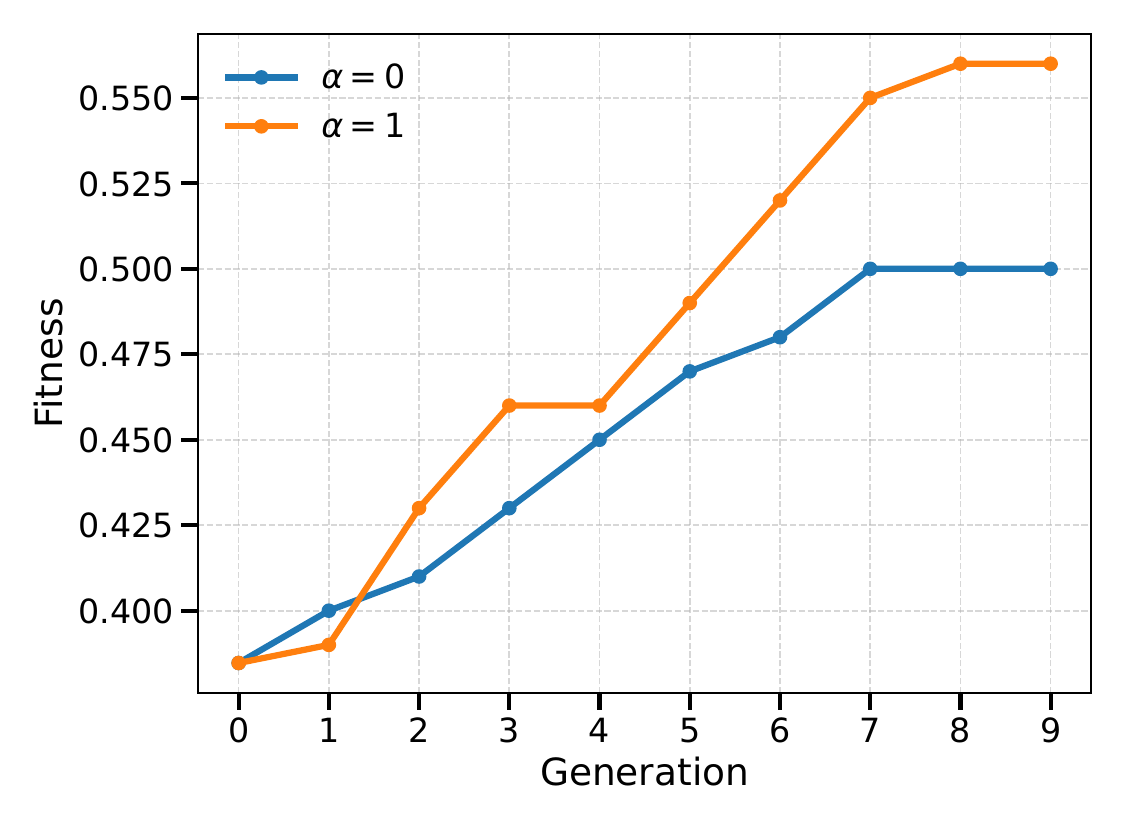}
\caption{Ablation of the Levenshtein regularization weight $\alpha$ in the evolutionary objective (Eq.~\ref{eq:levenstein}). Curves show best population fitness per generation for $\alpha=0$ and $\alpha=1$, averaged across evolutionary seeds.}
\label{fig:alpha_ablation}
\end{figure}

We study the effect of the Levenshtein regularization weight $\alpha$ (Eq.~\ref{eq:levenstein}) on evolutionary dynamics by comparing $\alpha=0$ (no structural regularization) and $\alpha=1$ (full regularization). As shown in Figure~\ref{fig:alpha_ablation}, enforcing structural similarity improves both convergence speed and final fitness relative to unconstrained mutation, indicating that similarity-aware variation stabilizes search in update-rule space. Across evolutionary runs, $\alpha=0$ and $\alpha=1$ reach maximum fitness values of approximately $0.50$ and $0.56$, respectively. In contrast, the intermediate setting $\alpha=0.5$, used for all main experiments reported in Figure~\ref{fig:evolution_curves}, consistently attains higher final fitness in the range $[0.65, 0.69]$ across both evolutionary seeds. This pattern indicates that partial structural regularization yields a more effective balance between preserving functional code structure and enabling exploratory variation.


\paragraph{Terminal value bootstrap.}
We ablate \textbf{CG-FPD} by adding a terminal value estimate to its planner. Specifically, we introduce a value head \(V(h)\) trained with one-step temporal-difference learning (TD(0)), i.e., regression to the immediate reward plus the estimated value of the next state, on the same robust-normalized reward used by \textbf{CG-FPD}. Its output is used only as a bounded terminal bonus in the planner score, and all other policy components and settings are identical to the baseline. Across all tested environments, adding a terminal-value bootstrap reduces peak performance relative to \textbf{CG-FPD} but consistently lowers variance across seeds, indicating improved stability. This suggests that \textbf{CG-FPD} is not limited by missing long-horizon return estimates; rather, introducing a learned value signal biases planning and degrades peak performance, even when used only as a bounded terminal bonus. The stability gains imply that value estimates act as a regularizer for planning, at the cost of restricting high-performing behavior.

\begin{table}[h]
\centering
\footnotesize
\setlength{\tabcolsep}{4pt}
\begin{tabular}{lcc}
\toprule
\textbf{Environment} &
\textbf{CG-FPD (baseline)} &
\textbf{CG-FPD + value bootstrap} \\
\midrule

\textbf{MountainCar-v0}
& \textbf{-105.80 \textpm\ 10.51}
& -122.12 \textpm\ 7.32 \\

\textbf{LunarLander-v3}
& \textbf{241.20 \textpm\ 11.00}
& 194.68 \textpm\ 7.04 \\

\textbf{Reacher}
& \textbf{-2.67 \textpm\ 0.15}
& -3.27 \textpm\ 0.11 \\

\bottomrule
\end{tabular}
\caption{Ablation of terminal value bootstrap added to \textbf{CG-FPD}. Results report mean $\pm$ standard deviation of peak evaluation return over random seeds.}
\label{tab:ablation_value_bootstrap}
\end{table}

\section{Limitation \& Future Work}
The framework is computationally expensive, since each candidate update rule must be evaluated through full reinforcement-learning training across multiple environments and random seeds, which restricts the scale of evolutionary search. In the current setting, the discovered algorithms arise from novel recombinations of existing reinforcement-learning mechanisms rather than from fundamentally new update forms, reflecting both the structure of the search space and the representational limits of the language model used to generate candidate rules.

Future work will, therefore, focus on scaling evolutionary search by improving evaluation efficiency and surrogate fitness models. As language models improve, we hypothesize that they will generate more complex and specialized update rules, potentially conditioned on task structure. An important direction is to explore richer combinations of canonical reinforcement-learning components with novel mechanisms generated by language models, and to study task-adaptive algorithm discovery.

\section{Conclusion}

In this work, we presented an evolutionary framework for discovering reinforcement-learning algorithms by searching directly over executable update rules. Building on REvolve, candidate training procedures are generated by large language models and selected based on empirical performance across environments, shifting optimization from architectures and hyperparameters to the learning procedure itself.

The evolved update rules achieve competitive performance across a diverse benchmark suite without relying on predefined actor-critic or temporal-difference structures. These results demonstrate that update-rule space is a viable target for evolutionary search and that language models can serve as effective generative operators over algorithmic code for automated reinforcement-learning algorithm discovery.

\begin{acks} 
This work is supported by Knut and Alice Wallenberg Foundation via the Wallenberg AI Autonomous Sensors Systems and the Wallenberg Scholars Grant.
We also acknowledge the National Academic Infrastructure for Supercomputing in Sweden (NAISS), partially funded by the Swedish Research Council through grant agreement no. 2022-06725, for awarding this project access to the LUMI supercomputer, owned by the EuroHPC Joint Undertaking and hosted by CSC (Finland) and the LUMI consortium.

\end{acks}

\bibliographystyle{ACM-Reference-Format}
\bibliography{references}

@inproceedings{revolve,
    title={{RE}volve: Reward Evolution with Large Language Models using Human Feedback},
    author={Rishi Hazra and Alkis Sygkounas and Andreas Persson and Amy Loutfi and Pedro Zuidberg Dos Martires},
    booktitle={The Thirteenth International Conference on Learning Representations},
    year={2025},
    url={https://openreview.net/forum?id=cJPUpL8mOw}
}

@inproceedings{henderson2018deep,
  title={Deep reinforcement learning that matters},
  author={Henderson, Peter and Islam, Riashat and Bachman, Philip and Pineau, Joelle and Precup, Doina and Meger, David},
  booktitle={Proceedings of the AAAI conference on artificial intelligence},
  volume={32},
  number={1},
  year={2018}
}

@article{engstrom2020implementation,
  title={Implementation matters in deep policy gradients: A case study on ppo and trpo},
  author={Engstrom, Logan and Ilyas, Andrew and Santurkar, Shibani and Tsipras, Dimitris and Janoos, Firdaus and Rudolph, Larry and Madry, Aleksander},
  journal={arXiv preprint arXiv:2005.12729},
  year={2020}
}

@article{patterson2024empirical,
  title={Empirical design in reinforcement learning},
  author={Patterson, Andrew and Neumann, Samuel and White, Martha and White, Adam},
  journal={Journal of Machine Learning Research},
  volume={25},
  number={318},
  pages={1--63},
  year={2024}
}

@article{franke2019neural,
  title={Neural architecture evolution in deep reinforcement learning for continuous control},
  author={Franke, J{\"o}rg KH and K{\"o}hler, Gregor and Awad, Noor and Hutter, Frank},
  journal={arXiv preprint arXiv:1910.12824},
  year={2019}
}

@article{jaderberg2017population,
  title={Population based training of neural networks},
  author={Jaderberg, Max and Dalibard, Valentin and Osindero, Simon and Czarnecki, Wojciech M and Donahue, Jeff and Razavi, Ali and Vinyals, Oriol and Green, Tim and Dunning, Iain and Simonyan, Karen and others},
  journal={arXiv preprint arXiv:1711.09846},
  year={2017}
}

@inproceedings{eimer2023hyperparameters,
  title={Hyperparameters in reinforcement learning and how to tune them},
  author={Eimer, Theresa and Lindauer, Marius and Raileanu, Roberta},
  booktitle={International conference on machine learning},
  pages={9104--9149},
  year={2023},
  organization={PMLR}
}

@inproceedings{pham2018efficient,
  title={Efficient neural architecture search via parameters sharing},
  author={Pham, Hieu and Guan, Melody and Zoph, Barret and Le, Quoc and Dean, Jeff},
  booktitle={International conference on machine learning},
  pages={4095--4104},
  year={2018},
  organization={PMLR}
}

@inproceedings{sato2021advantagenas,
  title={Advantagenas: Efficient neural architecture search with credit assignment},
  author={Sato, Rei and Sakuma, Jun and Akimoto, Youhei},
  booktitle={Proceedings of the AAAI Conference on Artificial Intelligence},
  volume={35},
  number={11},
  pages={9489--9496},
  year={2021}
}

@article{khadka2018evolution,
  title={Evolution-guided policy gradient in reinforcement learning},
  author={Khadka, Shauharda and Tumer, Kagan},
  journal={Advances in Neural Information Processing Systems},
  volume={31},
  year={2018}
}

@article{paul2019fast,
  title={Fast efficient hyperparameter tuning for policy gradient methods},
  author={Paul, Supratik and Kurin, Vitaly and Whiteson, Shimon},
  journal={Advances in Neural Information Processing Systems},
  volume={32},
  year={2019}
}

@article{xu2018meta,
  title={Meta-gradient reinforcement learning},
  author={Xu, Zhongwen and van Hasselt, Hado P and Silver, David},
  journal={Advances in neural information processing systems},
  volume={31},
  year={2018}
}

@article{he2022reinforcement,
  title={Reinforcement learning with automated auxiliary loss search},
  author={He, Tairan and Zhang, Yuge and Ren, Kan and Liu, Minghuan and Wang, Che and Zhang, Weinan and Yang, Yuqing and Li, Dongsheng},
  journal={Advances in neural information processing systems},
  volume={35},
  pages={1820--1834},
  year={2022}
}

@inproceedings{zou2021learning,
  title={Learning task-distribution reward shaping with meta-learning},
  author={Zou, Haosheng and Ren, Tongzheng and Yan, Dong and Su, Hang and Zhu, Jun},
  booktitle={Proceedings of the AAAI Conference on Artificial Intelligence},
  volume={35},
  number={12},
  pages={11210--11218},
  year={2021}
}

@inproceedings{zoph2017nas,
  title     = {Neural Architecture Search with Reinforcement Learning},
  author    = {Barret Zoph and Quoc V. Le},
  booktitle = {5th International Conference on Learning Representations (ICLR)},
  year      = {2017},
  note      = {Oral Presentation},
  url       = {https://openreview.net/forum?id=r1Ue8Hcxg}
}

@inproceedings{coreyes2021evolving,
  title     = {Evolving Reinforcement Learning Algorithms},
  author    = {John D. Co-Reyes and Xue Bin Peng and Sergey Levine and Pieter Abbeel and John Schulman},
  booktitle = {9th International Conference on Learning Representations (ICLR)},
  year      = {2021},
  url       = {https://openreview.net/forum?id=9XlAMdLMrB}
}

@inproceedings{oh2020discovering,
  title     = {Discovering Reinforcement Learning Algorithms},
  author    = {Junhyuk Oh and Rishabh Agarwal and Kelvin Xu and Dale Schuurmans and Quoc V. Le and Mohammad Norouzi},
  booktitle = {Advances in Neural Information Processing Systems (NeurIPS)},
  volume    = {33},
  pages     = {1060--1070},
  year      = {2020},
  url       = {https://papers.nips.cc/paper_files/paper/2020/hash/a322852ce0df73e204b7dfbce9c8cfd0-Abstract.html}
}

@inproceedings{eureka,
title={Eureka: Human-Level Reward Design via Coding Large Language Models},
author={Yecheng Jason Ma and William Liang and Guanzhi Wang and De-An Huang and Osbert Bastani and Dinesh Jayaraman and Yuke Zhu and Linxi Fan and Anima Anandkumar},
booktitle={The Twelfth International Conference on Learning Representations},
year={2024},
url={https://openreview.net/forum?id=IEduRUO55F}
}

@article{island_ea_seminal,
  title={A survey of parallel genetic algorithms},
  author={Cant{\'u}-Paz, Erick and others},
  journal={Calculateurs paralleles, reseaux et systems repartis},
  volume={10},
  number={2},
  pages={141--171},
  year={1998},
  publisher={Citeseer}
}

@article{stanley2002evolving,
  title={Evolving neural networks through augmenting topologies},
  author={Stanley, Kenneth O and Miikkulainen, Risto},
  journal={Evolutionary computation},
  volume={10},
  number={2},
  pages={99--127},
  year={2002},
  publisher={MIT Press}
}

@inproceedings{real2017large,
  title={Large-scale evolution of image classifiers},
  author={Real, Esteban and Moore, Sherry and Selle, Andrew and Saxena, Saurabh and Suematsu, Yutaka Leon and Tan, Jie and Le, Quoc V and Kurakin, Alexey},
  booktitle={International conference on machine learning},
  pages={2902--2911},
  year={2017},
  organization={PMLR}
}

@inproceedings{lcvenshtcin1966binary,
  title={Binary coors capable or ‘correcting deletions, insertions, and reversals},
  author={Lcvenshtcin, VI},
  booktitle={Soviet physics-doklady},
  volume={10},
  number={8},
  year={1966}
}

@inproceedings{lehman2011evolving,
  title={Evolving a diversity of virtual creatures through novelty search and local competition},
  author={Lehman, Joel and Stanley, Kenneth O},
  booktitle={Proceedings of the 13th annual conference on Genetic and evolutionary computation},
  pages={211--218},
  year={2011}
}

@book{Koza1992,
  title     = {Genetic Programming: On the Programming of Computers by Means of Natural Selection},
  author    = {Koza, John R.},
  year      = {1992},
  publisher = {MIT Press},
  address   = {Cambridge, MA, USA},
  isbn      = {9780262111706}
}

@misc{gym_mountaincar_wiki,
  author = {{OpenAI Gym}},
  title = {MountainCar-v0},
  howpublished = {\url{https://github.com/openai/gym/wiki/mountaincar-v0}},
  year = {2016},
  note = {OpenAI Gym Wiki}
}

@misc{gymnasium_general,
  author = {{Farama Foundation}},
  title = {Gymnasium: A Standard API for Reinforcement Learning},
  howpublished = {\url{https://gymnasium.farama.org/}},
  year = {2023},
  note = {Gymnasium documentation}
}

@misc{gym_leaderboard_lunar_lander,
  author = {{OpenAI Gym}},
  title = {Leaderboard Solved Definitions (LunarLander-v2)},
  howpublished = {\url{https://github.com/openai/gym/wiki/Leaderboard}},
  year = {2016},
  note = {Lists average reward 200 as solved for LunarLander}
}

@article{halfcheetah_benchmark_gac,
  author = {Hazim Alzorgan and Abolfazl Razi},
  title = {Monte Carlo Beam Search for Actor-Critic Reinforcement Learning in Continuous Control},
  journal = {arXiv preprint},
  year = {2025},
  note = {Reports HalfCheetah-v4 returns around 12750 for strong baselines},
  url = {https://arxiv.org/abs/2505.09029v1}
}

@inproceedings{real2019regularized,
  title={Regularized evolution for image classifier architecture search},
  author={Real, Esteban and Aggarwal, Alok and Huang, Yanping and Le, Quoc V},
  booktitle={Proceedings of the aaai conference on artificial intelligence},
  volume={33},
  number={01},
  pages={4780--4789},
  year={2019}
}

@article{salimans2017evolution,
  title={Evolution strategies as a scalable alternative to reinforcement learning},
  author={Salimans, Tim and Ho, Jonathan and Chen, Xi and Sidor, Szymon and Sutskever, Ilya},
  journal={arXiv preprint arXiv:1703.03864},
  year={2017}
}

@article{wang2019paired,
  title={Paired open-ended trailblazer (poet): Endlessly generating increasingly complex and diverse learning environments and their solutions},
  author={Wang, Rui and Lehman, Joel and Clune, Jeff and Stanley, Kenneth O},
  journal={arXiv preprint arXiv:1901.01753},
  year={2019}
}

@article{hemberg2024evolving,
  title={Evolving code with a large language model},
  author={Hemberg, Erik and Moskal, Stephen and O’Reilly, Una-May},
  journal={Genetic Programming and Evolvable Machines},
  volume={25},
  number={2},
  pages={21},
  year={2024},
  publisher={Springer}
}

@article{andrychowicz2016learning,
  title={Learning to learn by gradient descent by gradient descent},
  author={Andrychowicz, Marcin and Denil, Misha and Gomez, Sergio and Hoffman, Matthew W and Pfau, David and Schaul, Tom and Shillingford, Brendan and De Freitas, Nando},
  journal={Advances in neural information processing systems},
  volume={29},
  year={2016}
}

@inproceedings{nasir2024llmatic,
  title={Llmatic: neural architecture search via large language models and quality diversity optimization},
  author={Nasir, Muhammad Umair and Earle, Sam and Togelius, Julian and James, Steven and Cleghorn, Christopher},
  booktitle={proceedings of the Genetic and Evolutionary Computation Conference},
  pages={1110--1118},
  year={2024}
}

@article{konda1999actor,
  title={Actor-critic algorithms},
  author={Konda, Vijay and Tsitsiklis, John},
  journal={Advances in neural information processing systems},
  volume={12},
  year={1999}
}

@article{sutton1999policy,
  title={Policy gradient methods for reinforcement learning with function approximation},
  author={Sutton, Richard S and McAllester, David and Singh, Satinder and Mansour, Yishay},
  journal={Advances in neural information processing systems},
  volume={12},
  year={1999}
}

@article{jaderberg2016reinforcement,
  title={Reinforcement learning with unsupervised auxiliary tasks},
  author={Jaderberg, Max and Mnih, Volodymyr and Czarnecki, Wojciech Marian and Schaul, Tom and Leibo, Joel Z and Silver, David and Kavukcuoglu, Koray},
  journal={arXiv preprint arXiv:1611.05397},
  year={2016}
}

\clearpage
\appendix
\section*{Appendix Overview}
This appendix is organized as follows.  
Appendix~A provides detailed environmental specifications, reward definitions, and fitness normalization.  
Appendix~B describes the two best-evolved algorithms, \textbf{CG-FPD} and DF-CWP-CP, including their internal mechanisms, and presents the terminal-value bootstrap ablation for \textbf{CG-FPD} and its implementation details. 
Appendix~C reports the prompts used for the models.
\section{Environment Details}\label{app:env details}
\subsection{Fair Comparison and Fixed Policy Trunk}\label{app: fair comparison}

To isolate the effect of the learning update rule, we control representational capacity and optimization across all methods. Our PPO and SAC baselines use the standard MLP policy architecture with two 256-unit hidden layers, and we adopt the same policy trunk and action head dimensions for all discovered algorithms. Concretely, the policy network is fixed to a two-layer $256 \times 256$ Tanh MLP with a linear output head, and is trained with the same optimizer (Adam, $\mathrm{lr}=3\times 10^{-4}$). In the algorithm-discovery setting, candidate methods are therefore constrained to modify only the internal learning logic (e.g., the loss and auxiliary objectives) while keeping the policy parameterization and optimizer identical. This design ensures that performance differences are attributable to the learned update rule rather than to network size, architecture choice, or optimizer tuning. All discovered algorithms and baselines are evaluated every 5000 steps.

\begin{table}[h]
\centering
\renewcommand{\arraystretch}{1.2}
\setlength{\tabcolsep}{4pt}
\small
\begin{tabular}{|l|c|c|c|}
\hline
\textbf{Environment} 
& \textbf{State Dim.} 
& \textbf{Action Space} 
& \textbf{Action Type} \\
\hline
CartPole-v1      
& 4  
& $\{0,1\}$ 
& Discrete \\
\hline
MountainCar-v0  
& 2  
& $\{0,1,2\}$ 
& Discrete \\
\hline
LunarLander-v3  
& 8  
& $\{0,1,2,3\}$ 
& Discrete \\
\hline
Acrobot-v1      
& 6  
& $\{0,1,2\}$ 
& Discrete \\
\hline
InvertedPendulum-v4
& 4  
& $\mathbb{R}^1$ 
& Continuous \\
\hline
HalfCheetah-v5  
& 17 
& $\mathbb{R}^6$ 
& Continuous \\
\hline
Hopper-v4       
& 11 
& $\mathbb{R}^3$ 
& Continuous \\
\hline
Walker2d-v4     
& 17 
& $\mathbb{R}^6$ 
& Continuous \\
\hline
Reacher-v4      
& 11 
& $\mathbb{R}^2$ 
& Continuous \\
\hline
Swimmer-v4      
& 8  
& $\mathbb{R}^2$ 
& Continuous \\
\hline
\end{tabular}
\caption{Observation dimensionality and action space characteristics of the Gymnasium environments.}
\label{tab:env-spaces-detailed}
\end{table}

\begin{table}[h]
\centering
\renewcommand{\arraystretch}{1.2}
\setlength{\tabcolsep}{4pt}
\small
\begin{tabular}{|l|p{0.65\columnwidth}|}
\hline
\textbf{Environment} & \textbf{Reward} \\
\hline
CartPole-v1
& $+1$ per timestep the pole remains upright. \\
\hline
MountainCar-v0
& $-1$ per timestep until the goal is reached. \\
\hline
LunarLander-v3
& Shaped reward on position, velocity, and angle; fuel penalties; $+100$ for landing, $-100$ for crash. \\
\hline
Acrobot-v1
& $-1$ per timestep until the upright terminal state is reached. \\
\hline
InvPendulum-v4
& $+1$ per timestep while upright; termination on instability. \\
\hline
HalfCheetah-v5
& Forward velocity minus control cost. \\
\hline
Hopper-v4
& Forward velocity minus control cost plus alive bonus ($+1$). \\
\hline
Walker2d-v4
& Forward velocity minus control cost plus alive bonus ($+1$). \\
\hline
Reacher-v4
& Negative distance to target minus control cost (max reward $=0$). \\
\hline
Swimmer-v4
& Forward velocity minus control cost. \\
\hline
\end{tabular}
\caption{Reward definitions of the Gymnasium benchmark environments.}
\label{tab:env-rewards-detailed}
\end{table}

\subsection{Fitness Computation and Normalization}
\label{sec:appendix_fitness}

This section describes how scalar fitness values are computed from raw evaluation rewards during evolutionary search.

\paragraph{Per-seed evaluation.}
For a fixed update rule and environment, training is executed independently across multiple random seeds.
During training, evaluation episodes are periodically run and the resulting episodic returns are logged.
For each seed~$s$, we record the maximum evaluation return achieved during training,
\[
R_{\max}^{(s)} = \max_{t \le T} R_t^{(s)},
\]
where $R_t^{(s)}$ denotes the evaluation return at evaluation step~$t$.

\paragraph{Environment-level aggregation.}
For each environment~$i$, we aggregate performance across seeds by averaging the per-seed maxima,
\[
\bar{R}_i = \frac{1}{S} \sum_{s=1}^{S} R_{\max}^{(s)}.
\]
This metric captures the best performance an update rule is capable of achieving under each environment, independent of learning speed or intermediate instability.

\paragraph{Normalization to unit-scale fitness.}
Because environments differ substantially in reward scale and difficulty, raw returns are not directly comparable.
We therefore normalize environment-level scores using fixed reference bounds $(L_i, U_i)$:
\[
F_i = \mathrm{clip}\!\left(
\frac{\bar{R}_i - L_i}{U_i - L_i},\; 0,\; 1
\right).
\]
Here, $L_i$ and $U_i$ denote lower and upper reference rewards for environment~$i$.
These bounds are chosen based on known task structure (e.g., per-step penalties or terminal rewards)
and empirical calibration runs, and are held fixed across all evolutionary experiments.

\paragraph{Overall fitness.}
The final scalar fitness used for evolutionary selection is computed as the mean normalized score across environments,
\[
F = \frac{1}{N} \sum_{i=1}^{N} F_i.
\]
Only this scalar fitness is exposed to the evolutionary process; no gradient information or intermediate training signals are used.

\subsection{Training Environment Fitness}
\label{sec:training_env_fitness}

All training and evaluation environments are drawn from the Gymnasium framework, which provides a standardized and widely adopted API for reinforcement learning benchmarks \cite{gymnasium_general}. Because environments differ substantially in reward scale, episode structure, and termination conditions, raw evaluation returns are not directly comparable across tasks. We therefore normalize performance using environment-specific reference values derived from official task specifications and established benchmarking conventions.

For MountainCar-v0, the environment assigns a reward of $-1$ at every timestep until the agent reaches the goal position or the episode terminates after 200 steps. This yields feasible episodic returns in approximately $[-200,0]$. In practice, the OpenAI Gym benchmark defines MountainCar-v0 as solved when an agent achieves an average return of at least $-110$ over 100 consecutive episodes \cite{gym_mountaincar_wiki}. We treat this solved threshold as a meaningful upper reference point for normalization, rather than assuming the theoretical upper envelope of zero reward is achievable in practice.

LunarLander-v3) uses a shaped reward function based on lander position, velocity, orientation, fuel consumption, and terminal outcomes. While the environment specification does not define strict theoretical minimum or maximum returns, community benchmarks and the OpenAI Gym leaderboard conventionally regard an average return of $200$ or higher as indicative of successful task completion \cite{gym_leaderboard_lunar_lander}. Accordingly, performance below this threshold is not considered solved, and normalized fitness values reflect relative progress toward this benchmark rather than binary success.

For the HalfCheetah-v5 MuJoCo continuous control task, the reward is defined as forward velocity minus a control cost. Gymnasium does not specify a solved threshold or an explicit upper bound on achievable return for this environment. Instead, performance is typically interpreted relative to empirical benchmarks. In widely used open-source implementations and benchmark reports, standard actor–critic methods such as SAC and PPO achieve mean episodic returns in the range of several thousand under typical training budgets, with higher returns attainable under extended training and favorable hyperparameter settings \cite{halfcheetah_benchmark_gac}. We use these empirically observed scales solely to define an upper reference range for normalized fitness and not as a criterion for task completion or optimality.

\subsection{Metrics Feedback}\label{app:metrics feedback}
\paragraph{Training Metrics.}
For each update rule $f$, training produces a sequence of losses $\{\ell_t(f)\}$, gradient norms $\{\|\nabla_\theta \ell_t(f)\|\}$, and parameter norms $\{\|\theta_t\|\}$, where $\theta_t$ denotes the policy parameters at update step $t$. The loss $\ell_t(f)$ quantifies the magnitude and stability of the learning signal induced by the update rule. The gradient norm $\|\nabla_\theta \ell_t(f)\|$ serves as an indicator of numerical stability, with large values signaling exploding gradients and near-zero values indicating vanishing updates. The parameter norm $\|\theta_t\|$ captures long-term drift in the policy parameters and allows detection of pathological growth or collapse. Behavioral performance is measured separately by the mean evaluation return $J_t = \mathbb{E}[R \mid \pi_{\theta_t}]$, computed from periodic deterministic rollouts. Together, these quantities characterize optimization dynamics and policy quality without assuming any particular value-based or policy-gradient formulation.

\section{Best Evolved Algorithms}\label{app:best_evolved_algorithms}

\subsection{Cross-Entropy Guided Fractal Plan Distillation {CG-FPD}}
\label{cg-fpd}

The \textbf{CG-FPD} is a model-based reinforcement learning method that trains a policy through
\emph{plan distillation}.
Learning is driven by a learned latent world model and a planning-based teacher, while
execution at both training and evaluation time is performed by a single feedforward policy.

The method maintains the following learned components:
\begin{itemize}
    \item a policy network $\pi_\theta(a \mid s)$,
    \item a latent dynamics model $f_\phi(z_t, a_t) \rightarrow z_{t+1}$,
    \item auxiliary reward and termination predictors,
    \item an exponential moving average (EMA) target policy used for representation stabilization.
\end{itemize}

Notably, the algorithm does not learn a value function, Q-function, or advantage estimator.

\subsubsection{Latent Representation and Dynamics Learning}

Observations are processed by the policy network into latent feature vectors.
A latent dynamics model predicts the next latent state given the current latent state and an action:
\[
z_{t+1} = f_\phi(z_t, a_t).
\]

The dynamics model is trained using a combination of:
(i) regression toward target-policy latent features,
(ii) reward-aligned feature shaping, and
(iii) contrastive stabilization losses.
These objectives encourage the latent space to be both predictive and behaviorally relevant,
without explicitly modeling full environment state or transition probabilities.

\subsubsection{Planning as a Teacher Signal}

The central discovered mechanism is a planning-based teacher used exclusively for supervision, referred to as \emph{Cross-Entropy Guided Fractal Plan Distillation (CG-FPD)}.

At each policy update, the algorithm performs the following steps:

\begin{enumerate}
    \item \textbf{Sequence Sampling.}
    A set of $C$ candidate action sequences of fixed horizon $H$ is sampled from a proposal
    distribution centered on the current policy output.
    \item \textbf{Latent Rollout.}
    Each sequence is rolled forward in latent space using the learned dynamics model.
    \item \textbf{Sequence Scoring.}
    Each imagined trajectory is assigned a scalar score based on predicted reward,
    termination likelihood, and latent consistency penalties.
    \item \textbf{CEM Refinement.}
    A small number of Cross-Entropy Method (CEM) iterations refines the sequence distribution
    toward high-scoring regions.
    \item \textbf{Teacher Construction.}
    A teacher signal is formed from a weighted aggregation of the \emph{first actions}
    of the refined sequences.
\end{enumerate}

Formally, for a candidate action sequence
$\mathbf{a}_{0:H-1} = (a_0,\dots,a_{H-1})$, the planner assigns a score:
\[
S(\mathbf{a}_{0:H-1})
= \sum_{t=0}^{H-1}
\Big[
\gamma_t \, \hat r(z_t, a_t)
+ \lambda_s (1 - \hat d(z_t, a_t))
\Big]
- \lambda_c \, \mathcal{C}(z_{0:H}),
\]
where $\hat r$ and $\hat d$ denote learned reward and termination predictors, $\mathcal{C}$ is a latent consistency penalty, and $\gamma_t$ represents the survival weighting induced by predicted termination.

Only the first action of each planned sequence is retained.
This design preserves multi-step foresight while limiting the impact of compounding model error.

After CEM refinement, the teacher for state $s$ is defined using only the first action:
\[
a^{\text{teach}}_0
= \sum_{i=1}^{C} w_i \, a^{(i)}_0,
\qquad
w_i = \frac{\exp(S_i / \tau)}{\sum_j \exp(S_j / \tau)},
\]
where $S_i$ is the score of sequence $i$ and $\tau$ is a temperature parameter.

\subsubsection{Policy Update via Distillation}

The policy is trained to match the teacher signal produced by CG-FPD.
For discrete action spaces, the teacher defines a target action distribution, and the policy is updated via cross-entropy.
For continuous action spaces, the teacher is a target action vector, and the policy is trained via regression with additional regularization.

The resulting planning loss is
\[
\mathcal{L}_{\text{plan}}
=
\begin{cases}
\mathrm{CE}\!\left(\pi_\theta(\cdot \mid s), \, p^{\text{teach}}(\cdot)\right),
& \text{(discrete)} \\[4pt]
\left\lVert \pi_\theta(s) - a^{\text{teach}}_0 \right\rVert_1,
& \text{(continuous)},
\end{cases}
\]
and is optimized jointly with auxiliary smoothness, diversity, anchoring,
and representation-learning losses.
All updates are performed via standard backpropagation, without computing returns,
advantages, or policy gradients.

\subsubsection{Relation to Existing Methods}

While individual components of the algorithm—latent dynamics models, sequence-based planning, and policy distillation—are present in prior work, their combination in this configuration is uncommon. In particular, the exclusive use of planning as a supervised teacher, combined with the absence of value functions or policy gradients, distinguishes the discovered algorithm from existing model-based and model-free approaches.

\subsubsection{Summary}

In summary, the discovered algorithm:
\begin{itemize}
    \item learns a policy without value functions or policy gradients,
    \item uses latent-space planning exclusively as a supervised teacher signal,
    \item distills only the first action of short-horizon imagined trajectories,
    \item and was obtained through automated evolutionary search guided by a large language model.
\end{itemize}

The complete optimization objective is implemented in the function
\texttt{compute\_loss}, which aggregates latent dynamics learning,
planning-based distillation, and auxiliary stabilization losses.

\subsection{Dual-Flow Confidence World Planning with Controllability Prior (DF-CWP-CP)}
\label{app:df-cwp-cp}

The DF-CWP-CP algorithm relies on short-horizon world-model planning, confidence-aware risk control, and latent policy-flow regularization to shape behavior.

\paragraph{World models and confidence.}
The algorithm learns differentiable predictors for forward dynamics, reward, and termination,
\[
\hat{f}_\phi(s_t, a_t), \quad \hat{r}_\phi(s_t, a_t, s_{t+1}), \quad \hat{d}_\phi(s_{t+1}),
\]
operating directly in observation space.
In parallel, confidence heads estimate the reliability of each prediction,
\[
c^{\text{dyn}}_t = \sigma(\hat{c}_{\text{dyn}}(s_t,a_t)), \quad
c^{\text{r}}_t = \sigma(\hat{c}_{\text{r}}(s_t,a_t,s_{t+1})), \quad
c^{\text{done}}_t = \sigma(\hat{c}_{\text{done}}(s_{t+1})).
\]
The combined confidence:
\[
c_t = c^{\text{dyn}}_t \, c^{\text{r}}_t \, c^{\text{done}}_t \in [0,1],
\]
is used during planning to down-weight low-confidence imagined transitions, explicitly discouraging exploitation of model error.

\paragraph{Dual-flow policy regularization.}
The policy produces bounded latent action codes:
\[
\tilde{a}_t = \tanh(\pi_\theta(s_t)).
\]
Two exponential moving average (EMA) copies of the policy, $\pi_{\theta_f}$ (fast) and $\pi_{\theta_s}$ (slow), define reference latent flows.
Training enforces alignment between the current latent temporal difference
\[
\Delta \tilde{a}_t = \tilde{a}_{t+1} - \tilde{a}_t
\]
and the corresponding EMA-induced flows $\Delta \tilde{a}^{(f)}_t$ and $\Delta \tilde{a}^{(s)}_t$ via cosine similarity penalties.
These constraints representational drift and stabilize long-horizon behavior without critics or value estimates.

\paragraph{Planning-based learning signal.}
At each update, the policy is optimized through differentiable short-horizon imagined rollouts of length $H$.
From an initial state $s_0$, the planner recursively computes
\[
s_{k+1} = s_k + \hat{f}_\phi(s_k, a_k),
\quad a_k = \tanh(\pi_\theta(s_k)).
\]
At each step, a bounded desirability score is accumulated,
\[
o_k = c_k \big(\hat{r}_\phi(s_k,a_k,s_{k+1}) - \lambda_d \hat{d}_\phi(s_{k+1}) \big)
      - \lambda_r (1 - c_k),
\]
where $\lambda_d$ penalizes termination and $\lambda_r$ penalizes low-confidence predictions.
The per-step score is saturated,
\[
\bar{o}_k = \tanh(o_k),
\]
and aggregated into a local planning objective:
\[
G = \frac{1}{H} \sum_{k=0}^{H-1} \gamma^k \bar{o}_k .
\]
This quantity is \emph{not} a return, value function, or advantage, but a short-horizon desirability optimized directly via backpropagation through the world model.

\paragraph{Controllability augmentation.}
To provide a learning signal in sparse or flat-reward regimes, the planner additionally maximizes a controllability objective defined as the action-sensitivity of the desirability,
\[
C = \frac{1}{H} \sum_{k=0}^{H-1} \gamma^k \,
c_k \left\| \frac{\partial \bar{o}_k}{\partial a_k} \right\|_2 .
\]
This term encourages actions that meaningfully influence predicted outcomes.
The gradient norm is clipped and regularized to prevent pathological sensitivity and is gated by model confidence to avoid amplifying unreliable predictions.

\paragraph{Anchor and regularization mechanisms.}
During imagined rollouts, actions are softly anchored to the slow EMA policy,
\[
\mathcal{L}_{\text{anchor}} =
\mathbb{E}\big[ c_k \| a_k - a^{(s)}_k \|_2^2 \big],
\]
reducing oscillations and instability when confidence is sufficient.
Additional regularizers—including entropy (for discrete actions), action-magnitude penalties, logit norms, and state-magnitude penalties—ensure bounded, conservative planning behavior.

\paragraph{Optimization.}
World-model losses, confidence calibration losses, latent flow regularization, auxiliary latent prediction, and the planning objective:
\[
\mathcal{L}_{\text{plan}} = -G - \lambda_c C + \text{regularizers},
\]
are jointly optimized using standard gradient descent.
The contribution of the planning objective is gradually increased via a warm-up schedule.
No Bellman equations, critics, value functions, advantage estimates, or policy-gradient objectives are used at any stage.

\subsubsection{Relation to Existing Methods}

The algorithm combines model-based learning, planning, and representation regularization in a distinct way. Unlike classical model-based reinforcement learning, learned world models are not used to form value functions, Bellman targets, or policy gradients, but only to define a local planning objective. Unlike model-predictive control, planning is used only during training as a supervision signal; execution relies on a single feedforward policy without online optimization. EMA policies resemble stabilization methods in representation learning, but here define reference temporal flows in latent action space rather than prediction targets. Overall, the method avoids value estimation and online planning while using short-horizon, confidence-aware planning to shape the policy.

\paragraph{Summary.}
Overall, the algorithm learns by repeatedly shaping the policy to be (i) consistent with learned world dynamics, (ii) stable in latent temporal flow, and (iii) locally optimal under confidence-aware, controllability-augmented planning. This results in a planning-driven learning framework that is distinct from both classical model-free reinforcement learning and standard model-predictive control.

\subsection{Ablation: Universal Terminal Value Bootstrap for CG-FPD}
\label{app:cg-fpd-value-bootstrap}

\textbf{Objective.} Test whether short-horizon planning in \textbf{CG-FPD} benefits from a bounded terminal value term without converting the method into an actor–critic.

\paragraph{Implementation delta (vs.\ base CG-FPD).}
\begin{itemize}
  \item \textbf{Added value head} $V(h)=\mathrm{MLP}(h)\!\to\!\mathbb{R}$ on policy features; optimized with TD(0) on the same robust-normalized reward used by the planner:
  \[
    L_V=\mathrm{Huber}\!\Big(V(s_t),~ \tilde r_t + \gamma(1-d_t)\,V(s_{t+1})\Big),\quad
    \tilde r_t=\tanh\!\Big(\frac{r_t-\mathrm{med}_r}{1.4826\,\mathrm{mad}_r+\varepsilon}\Big).
  \]
  \item \textbf{Planner score augmentation} at the terminal latent state $h_T$:
  \[
    S_{\text{new}} = S_{\text{CG-FPD}}
    ~+~ \lambda_{\mathrm{eff}} \,\mathrm{util}_{\max}\,
    \tanh\!\Big(\frac{V(h_T)-\mu_V}{\sigma_V+\varepsilon}\Big),
  \]
  where $(\mu_V,\sigma_V)$ are running global statistics of $V$, and $\mathrm{util}_{\max}$ is the existing planner utility clip.
  \item \textbf{Gating and safety.} No bootstrap during a warm-up of $50$k steps; enable as a function of TD-RMSE EMA:
  enable if $<0.5$, disable if $>0.8$, linear in-between, yielding $\lambda_{\mathrm{eff}}\in[0,\lambda]$.
  \item \textbf{Training/infra unchanged.} Policy head, optimizer, horizons, CEM settings, losses, and distillation remain identical; $V$ influences only the teacher, not the policy loss.
\end{itemize}

\paragraph{Default hyperparameters.}
$\gamma{=}0.99$, TD loss weight $=0.10$, $\lambda\in\{0.1,0.3,0.6\}$ (swept), RMSE EMA rate $=0.05$, value-stat EMA rate $=0.01$, warm-up $=50$k steps.

\paragraph{Reporting.}
Per environment: peak eval return (↑) and seed std (↓) for baseline vs.\ +bootstrap, $\Delta$ to baseline, median steps-to-peak (↓), and fraction of seeds within 95\% of best (“success rate”, ↑). Also log post–warm-up $\overline{\lambda_{\mathrm{eff}}}$ to confirm engagement.

\paragraph{Scope.}
This ablation keeps \textbf{CG-FPD} critic-free at the policy level; $V$ is auxiliary and used only for terminal bias in planning.
\clearpage
\section{Prompts}

\begin{promptbox}[label=system_prompt]{SYSTEM PROMPT}
\tiny
\begin{lstlisting}[language=Python, basicstyle=\tiny\ttfamily,
                   breaklines=true, breakatwhitespace=false, columns=fullflexible,
                   frame=none]
You are a research scientist designing a completely new reinforcement-learning algorithm
### Problem Context
An agent interacts with an environment over discrete time steps.

At each step t:
    s_t = current state (observation) provided by the environment  
    a_t = action chosen by the agent  
    s_{t+1}, r_t, done, info = env.step(a_t)

The environment defines:
- A state (observation) space S  
- An action space A  
- A transition function P(s_{t+1} | s_t, a_t)  
- A reward signal r_t (which may be re-interpreted by your algorithm)  
Termination:
- done is either True or False

Goal:
Implement a completely new reinforcement-learning algorithm.
For fair comparison, keep the same network layer sizes, the same action head, and the same optimizer settings as the baseline.
You may redesign all other internal logic.

Objective:
The algorithm must improve long-horizon task performance, measured by cumulative environment reward over episodes, using a mathematically novel update rule.

Research constraints:
Invent a new algorithm that is NOT based on:
- Bellman recursion or temporal-difference targets
- Q-learning, actor-critic, or policy-gradient methods
- Evolutionary optimization, cross-entropy methods, or imitation learning

Required:
Your algorithm must introduce an original internal learning mechanism that:
- Defines an update rule directly from (s_t, a_t, r_t, s_{t+1}, done)
- Updates model parameters to improve behavior
- Works in both discrete and continuous action spaces
- Avoids NaN/Inf and unstable numerics

- Can produce stable learning in both discrete and continuous action spaces,
- Optionally introduces internal normalization, adaptive signals, or auxiliary predictive modules,
- Defines its own differentiable or emergent learning signal that replaces the standard value/policy gradient paradigm.

---

### Environment Awareness
CartPole-v1 (discrete, obs.shape = (4,), actions = {0, 1})
MountainCar-v0 (discrete, obs.shape = (2,), actions = {0, 1, 2})
Acrobot-v1 (discrete, obs.shape = (6,), actions = {0, 1, 2})
HalfCheetah-v5 (continuous, obs.shape = (17,), action.shape = (6,), range = [-1.0, 1.0])
LunarLander-v3 (discrete, obs.shape = (8,), actions = {0, 1, 2, 3})

Therefore:
  - It must handle both **discrete** and **continuous** control.
  - It must generate valid actions for the respective spaces.
  - It must be general enough to run without manual code edits.
  - It must rely only on signals available from (s, a, r, s_next, done).

### Expected Output Behavior
Your algorithm should:
- Learn from environmental feedback even if the reward is sparse or delayed.
- Show interpretable adaptation over time (not random or unstable behavior).
- Produce measurable improvements in `eval/mean_reward` during training.
- Avoid unsafe numerical operations (NaN/inf in loss or gradients).

### Output Format
Write your proposal as a structured research concept:

**Algorithm Name:**  

**Core Idea:**  
Summarize the key intuition or mechanism behind your new update principle.

**Mathematical Intuition (high level):**  
Describe the internal computation flow or pseudo-loss driving learning.

Avoid any reference to existing RL algorithm names or terminology.  
Treat this as a conceptual research draft that will later be turned into a full Python implementation using a standardized base class.

### Integration Note
Your output from this stage will be combined automatically with an implementation prompt that transforms it into runnable Python code.
Ensure the concept you propose is **implementable using only standard PyTorch tensors** and functions available within the given framework.

Explain your reasoning clearly and analytically.


\end{lstlisting}
\end{promptbox}

\begin{promptbox}[label=STRUCTURAL MACRO-MUTATION]{STRUCTURAL MACRO-MUTATION}
\tiny
\begin{lstlisting}[language=Python, basicstyle=\tiny\ttfamily,
                   breaklines=true, breakatwhitespace=false, columns=fullflexible,
                   frame=none]
==============================
EVOLUTIONARY OPERATOR: STRUCTURAL MACRO-MUTATION
==============================

Your task is to apply a STRUCTURAL MACRO-MUTATION to the given algorithm.

This operator authorizes a substantial rewrite of EXACTLY one major internal
mechanisms of the algorithm but in a directed way:
the mutation must be guided by the observed performance in the metrics, fitness
summary, and any runtime errors.

Use these signals to identify which internal mechanisms are underperforming,
unstable, misaligned, or uninformative, and redesign the relevant internal
components to correct these weaknesses. Mutations must not be arbitrary; they
must be **justified by the failure patterns** visible in the provided results.

The external API, class name, method names, policy architecture, optimizer,
action-generation rules, and all refinement-stage restrictions remain unchanged.
Only the internal learning logic may be structurally rewritten.

---

### Input Algorithm
python '''
{Your_Algorithm}
'''

### Historical Metrics
{Metrics}

### Fitness Summary
{Fitness}

### Runtime Errors
{Errors}

---

### Output
Return the complete mutated of the algorithm as a full Python class named `NewAlgo`.
Computation Budget:
  All algorithms receive the same total number of environment steps (e.g., 1 M).
  The rollout_len parameter in the base code only defines how often data are collected before an update call.
  You may internally decide how to use that data (full-batch, online, replay, or streaming),
  but you must stay within the same total sample budget for fairness.
Rules:
- Keep the **same policy network** for action output (two 256-unit + activation layers + linear head) and the same optimizer (Adam, lr=3e-4).  
- Keep all import statements at the top of the file, outside the class. Never import inside the class or assign modules to `self` (e.g., `self.torch = torch` is forbidden).  
- Use decorators directly (e.g., `@torch.no_grad()`), never `@self.torch.no_grad()`.  
- Do not redesign exploration. You may adjust at most one scalar exploration-strength parameter for training (deterministic=False). When deterministic=True, evaluation must be fully deterministic.
- Keep the existing class and method names (`__init__`, `predict`, `learn`, `compute_loss`, `_update`), so the code remains compatible with the evaluation pipeline.  
- The  algorithm must prioritize stable training: no exploding exploration, no action-shape changes, and no redefinition of the policy network
- predict() must always return actions with correct shape: (act_dim,) for continuous or an integer for discrete. Do not add batch dimensions.
-However, the internal learning logic must not directly reproduce any known RL training formula such as policy-gradient, TD, or Bellman-style updates.  
-The algorithm should not rely on explicit advantage estimates, critic targets, or likelihood-ratio terms.  
-Reward may be used in any differentiable way but not inside any policy-gradient, TD, or Bellman-style update.  
-The method must remain distinct from all known RL algorithms (e.g., PPO, A2C, REINFORCE, TD-error, SAC, or Q-learning).  

Important:
You are free to define entirely new update rules or gradient signals, even if they deviate from conventional RL training, as long as they remain differentiable and trainable in PyTorch.  
You may combine or adapt principles from multiple paradigms to create a genuinely new mechanism that is not a copy or reimplementation of any existing RL algorithm.  
The algorithm must optimize behavior over long-horizon consequences rather than only immediate rewards. 
You are permitted to maintain temporally extended internal quantities that influence learning, as long as they do not compute or imitate returns, advantages, TD targets, Q-values, or critic functions.


Your output must be inside python ''' NewAlgo ''' and the name of the class must be exactly this NewAlgo.
Style of output:
python '''
<imports>
<class NewAlgo ...>
'''

\end{lstlisting}
\end{promptbox}

\begin{promptbox}[label=STRUCTURAL CROSSOVER]{STRUCTURAL CROSSOVER}
\tiny
\begin{lstlisting}[language=Python, basicstyle=\tiny\ttfamily,
                   breaklines=true, breakatwhitespace=false, columns=fullflexible,
                   frame=none]
==============================
EVOLUTIONARY OPERATOR: STRUCTURAL CROSSOVER
==============================

You are given the current reinforcement-learning algorithm below, plus evaluation results,
historical metric traces, fitness summaries, and any runtime errors.

Your task is to apply a **STRUCTURAL CROSSOVER**: synthesize a new algorithm by integrating
complementary internal components from **two parent algorithms** (the provided in-context samples).
You must identify which internal mechanisms of each parent are strong or weak based on their
reported metrics and fitness, and selectively recombine the strongest modules into a single
coherent learning system.

Crossover is **directed**, not random: differences in performance across parents must guide
which parts you preserve, replace, or merge. You may blend, unify, or re-architect internal components,
but the external API, class name, method names, policy architecture, optimizer, action rules, and
all refinement-stage restrictions must remain unchanged.

You must preserve:
  - the class name and all method names,
  - the fixed policy network architecture (two 256-unit + activation layers + linear head),
  - the optimizer and training budget,
  - differentiability and PyTorch compatibility,
  - the environment interface and action-generation rules,
  - the constraints that forbid standard RL methods (no policy gradients, no TD, no Q, no critics).

You may recombine or redesign internal mechanisms in compute_loss(), internal shaping signals,
latent transformations, or auxiliary modules. Your crossover must produce a **single** integrated
learning rule that forms a coherent algorithm rather than a concatenation of fragments.

Use the performance differences between the two parents (reflected in their metrics, fitness,
and errors) to decide:
  - which internal components to keep from parent A,
  - which components to inherit or modify from parent B,
  - which weak components to discard or rewrite.

---

### Input Algorithms
python '''
{Your_Algorithm}
'''

### Historical Metrics
{Metrics}

### Fitness Summary
{Fitness}

### Runtime Errors
{Errors}

---

### Output
Return the **complete crossover-derived version** of the algorithm as a full Python class named `NewAlgo`.




\end{lstlisting}
\end{promptbox}

\begin{promptbox}[label=STRUCTURAL CROSSOVER]{STRUCTURAL CROSSOVER (PART II)}
\tiny
\begin{lstlisting}[language=Python, basicstyle=\tiny\ttfamily, breaklines=true]
Computation Budget:
  All algorithms receive the same total number of environment steps (e.g., 1 M).
  The rollout_len parameter only defines how often data are collected before an update call.
  You may internally decide how to use that data (full-batch, online, replay, or streaming),
  but must remain within the same total sample budget.

Rules:
- Keep the **same policy network** for action output (two 256-unit + activation layers + linear head) and the **same optimizer** (Adam, lr=3e-4).    
- Keep all import statements at the **top of the file**, outside the class. Never import inside the class or assign modules to `self` (e.g., `self.torch = torch` is forbidden).  
- Use decorators directly (e.g., `@torch.no_grad()`), never `@self.torch.no_grad()`.  
- Do not redesign exploration. You may adjust at most one scalar exploration-strength parameter for training (deterministic=False). When deterministic=True, evaluation must be fully deterministic.
- Keep the existing class and method names (`__init__`, `predict`, `learn`, `compute_loss`, `_update`), so the code remains compatible with the evaluation pipeline.  
- The  algorithm must prioritize stable training: no exploding exploration, no action-shape changes, and no redefinition of the policy network
- predict() must always return actions with correct shape: (act_dim,) for continuous or an integer for discrete. Do not add batch dimensions.
-However, the internal learning logic must not directly reproduce any known RL training formula such as policy-gradient, TD, or Bellman-style updates.  
-The algorithm should not rely on explicit advantage estimates, critic targets, or likelihood-ratio terms.  
-Reward may be used in any differentiable way but not inside any policy-gradient, TD, or Bellman-style update.  
-The method must remain distinct from all known RL algorithms (e.g., PPO, A2C, REINFORCE, TD-error, SAC, or Q-learning).  

Important:
You are free to define entirely new update rules or gradient signals, even if they deviate from conventional RL training, as long as they remain differentiable and trainable in PyTorch.  
You may combine or adapt principles from multiple paradigms to create a genuinely new mechanism that is not a copy or reimplementation of any existing RL algorithm.  
The algorithm must optimize behavior over long-horizon consequences rather than only immediate rewards. It must incorporate internal reasoning that reflects multi-step outcomes without using returns, advantages, TD updates, Q-values, critics, or any value-based targets.
You are permitted to maintain temporally extended internal quantities that influence learning, as long as they do not compute or imitate returns, advantages, TD targets, Q-values, or critic functions.


Your output must be inside python ''' NewAlgo ''' and the name of the class must be exactly `NewAlgo`.
Style of output:
python '''
<imports>
<class NewAlgo ...>
'''
\end{lstlisting}
\end{promptbox}

\end{document}